\documentclass[10pt,journal,compsoc]{IEEEtran}

\usepackage{bm}
\usepackage{float}
\usepackage[pdftex]{graphicx}
\usepackage{graphicx}
\usepackage{amsmath}
\usepackage{amssymb}
\usepackage{cite}
\usepackage{url}
\usepackage{color}
\usepackage{subfig}
\usepackage{ragged2e}
\usepackage{tikz}
\usepackage{tabularx}
\usepackage{makecell}
\usetikzlibrary{chains, positioning, trees}
\usepackage[utf8]{inputenc}

\usepackage{algorithm}
\usepackage{algorithmic}
\usepackage{multirow}
\usepackage{ulem}

\tikzset{
conv/.style={draw,circle,fill=red!20,text width=0.6cm,text centered},
act/.style={draw,rectangle,fill=blue!20,text width=1cm,text centered},
pool/.style={draw,rectangle,fill=green!20,text width=1cm,text centered},
}

\begin{document}

\title{The Expressive Power of Graph Neural Networks: A Survey}

\author{Bingxu Zhang*,
        Changjun Fan*,
        Shixuan Liu,
        Kuihua Huang,
        Xiang Zhao,
        Jincai Huang,
        and Zhong Liu
    
\IEEEcompsocitemizethanks{\IEEEcompsocthanksitem B. Zhang, C. Fan, S. Liu, K. Huang, X. Zhao, J. Huang and Z. Liu are with the Laboratory for Big Data and Decision, College of Systems Engineering, National University of Defense Technology, Hunan, China. Email: \{zhangbingxunudt, fanchangjun, liushixuan, huangkuihua, zhaoxiang, huangjincai, liuzhong\}@nudt.edu.cn}
\thanks{*These authors contributed equally.}
\thanks{(Corresponding authors: Changjun Fan, Zhong Liu.)}
}


\markboth{IEEE TRANSACTIONS ON KNOWLEDGE AND DATA ENGINEERING}%
{Shell \MakeLowercase{\textit{et al.}}: The Expressive Power of Graph Neural Networks: A Survey}

\maketitle

\begin{abstract}
\justifying
Graph neural networks (GNNs) are effective machine learning models for many graph-related applications. Despite their empirical success, many research efforts focus on the theoretical limitations of GNNs, i.e., the GNNs expressive power. Early works in this domain mainly focus on studying the graph isomorphism recognition ability of GNNs, and recent works try to leverage the properties such as subgraph counting and connectivity learning to characterize the expressive power of GNNs, which are more practical and closer to real-world. However, no survey papers and open-source repositories comprehensively summarize and discuss models in this important direction. To fill the gap, we conduct a first survey for models for enhancing expressive power under different forms of definition. Concretely, the models are reviewed based on three categories, i.e., Graph feature enhancement, Graph topology enhancement, and GNNs architecture enhancement.

\end{abstract}


\maketitle

\section{Introduction}\label{sec1}

\IEEEPARstart{G}{raph} neural networks (GNNs) have emerged as a leading model in deep learning, attracting significant research interests~\cite{hamilton2017inductive, kipf2018neural, velivckovic2017graph} due to their superior ability in learning graph data. Their variants are widely applied in various real-world scenarios, including \cite{wang2019kgat}, computer vision\cite{yang2019auto}, natural language processing\cite{chen2019reinforcement}, molecular analysis\cite{ma2018constrained}, data mining\cite{leclair2020improved}, financial risk control~\cite{tian2023asa, zhou2020graph}, bioinformatics\cite{wang2022progressive} and anomaly detection\cite{markovitz2020graph}. For more introductions to the foundations and applications of GNNs, see references~\cite{zhang2020deep, wu2020comprehensive, zhou2020graph, Liu2021SamplingMF, 9875203, 10266657, liu2024inductive} for more details.

Advancements in GNNs have garnered academic interest and hold significant potential for real-world applications, offering practical solutions to complex issues. For instance, in community detection\cite{zhu2021unsupervised}, GNNs leverage mutual information maximization to uncover structures within attribute networks. In security domain\cite{hou2023sequential}, sequential attention methods utilizing GNNs feature representations are adept at tracking network dynamic changes. Recommendation systems\cite{cai2023expressive} also benefit from the GNNs rich nodes representation, adeptly handling extensive user behavior and item attribute data to enhance recommendation accuracy. Overall, GNNs is pivotal across various domains, enhancing task performance and pioneering novel approaches to problem-solving. This versatility cannot be achieved without the support of the powerful expressive power of GNNs, which is essential for effectively capturing and modelling the complex dynamics of real-world data.

The expressive power of neural networks (NNs) refers to their ability to represent and approximate a wide range of functions, patterns, and relationships within data. It encompasses the network's capacity to capture and model complex and intricate relationships present in the input data. NNs are known for their remarkable expressive power, which allows them to learn and represent highly non-linear and intricate relationships between input and output data. While the expressive power of GNNs refers to the scope and complexity that the model is able to express when processing graph data, specifically the model's ability to encode and process feature information on the graph as well as graph structure information. Suppose we have a social network dataset and the goal is to identify users with high influence in the network, where the authority of the user (node features) and the social scope of the user (neighborhood structure of the nodes) together determine the influence of the user. Using GNNs to capture the relationships between nodes and the topology of the network, by aggregating and transforming the node features and learning the complex patterns of the network structure, it is possible to generate node representations with rich expressive power that reflect the feature and structural relationships of the nodes, thus helping us to identify high-influence users. With this example, it can be seen that the expressive power of GNNs plays an important role when dealing with graph data by allowing us to capture the complex patterns of the network with rich representations that guide our reasoning and decision-making process. Thus, expressive power is one of the core concepts of GNNs and provides us with a powerful tool for understanding and processing graph data.

However, studying the expressive power of GNNs faces many challenges. Firstly, Compared to well-structured texts and images, graphs are irregular. A fundamental assumption behind machine learning on graphs is that the predicted targets should be invariant to the order of nodes on graphs. To match this assumption, GNNs introduce an inductive bias called permutation invariance~\cite{battaglia2018relational}. Specifically, GNNs give outputs that are independent of how the node indices of the graph are assigned and the order in which they are processed, i.e., the model parameters are independent of the node order and are shared throughout the graph. Due to this new parameter sharing mechanism, we need new theoretical tools to study their expressive power. Moreover, most GNNs are usually used as black-box feature extractors for graphs, and it is unclear how well they can capture different graph features and topology. Secondly, due to the introduction of permutation invariance, the results of the classical universal approximation theorem of NNs ~\cite{cybenko1992approximation, hornik1991approximation} cannot be directly generalized to GNNs~\cite{maron2019universality,azizian2020expressive,keriven2019universal}. In addition, in practice, the study of expressive power is related to a number of long-standing difficult problems in graph theory~\cite{helfgott2017graph, hartmanis1982computers}. For example, in the prediction of properties of chemical molecules, it is necessary to judge whether the molecular structure is the same or similar to that of a molecule with known properties, which involves the problem of graph/subgraph isomorphism judgement~\cite{helfgott2017graph, levin2000stochastic} and graph matching~\cite{bunke2000graph, zanfir2018deep, lewis1983michael}, etc.

There exists a pioneering study of the expressive power of GNNs. Morris et al.~\cite{morris2019weisfeiler,morris2020weisfeiler} and Xu et al.~\cite{xu2018powerful} introduced to use graph isomorphism recognition to explain the expressive power of GNNs, leading the trend to analyze the separation ability of GNNS. Maron et al.~\cite{maron2019universality} and Chen et al.~\cite{chen2019equivalence} introduced to use the ability of GNNs to approximate graph functions to account for their expressive power and further gave the set representation of invariant graph functions that can be approximated by GNNs, leading the trend to analyze the approximation ability of GNNs. While multiple studies have emerged in recent years that characterize and enhance the expressive power of GNNs, a comprehensive review in this direction is still lacking. Sato~\cite{sato2020survey} explored the relationship between graph isomorphism testing and expressive power and summarized strategies to overcome the limitations of GNNs expressive power. However, they only describe the GNNs separation ability and approximation ability to characterize the GNNs expressive power, and there also exists other abilities, including subgraph counting ability, spectral decomposition ability~\cite{zhu2021interpreting, balcilar2021analyzing, kanatsoulis2022graph}, logical ability~\cite{de2000note, cai1992optimal, geerts2022expressive, balcilar2021breaking, grohe2021logic, barcelo2020expressive, barcelo2020logical}, etc., which are also considered as the main categories of GNNs expressive power. Therefore, it is crucial to clarify the definition and scope of the GNNs expressive power, which motivates this work.

In this work, we explore GNNs expressive power through two critical components: feature embedding and topology representation. The latter capability distinguishes GNNs within other machine learning models. Our analysis reveals that both features and topology influence GNNs expressive power, with topology learning and keeping deficiencies being a major limitation.

This paper categorizes methods to enhance GNNs expressive power into three main strategies: graph feature enhancement, which boosts embedding effects; graph topology enhancement, aimed at more effectively representing complex topological information; and GNNs architecture enhancement, focusing on refining aggregation functions that restrict expressivity.

We spotlight the deficiencies in current benchmarks and metrics for GNNs expressive power, underscoring the obstacles in gauging it. We also propose innovative research trajectories, such as physics-informed GNN design and the exploration of alternative graph transformer models.

\begin{figure}[!t]
\centering
\includegraphics[width=0.5\textwidth]{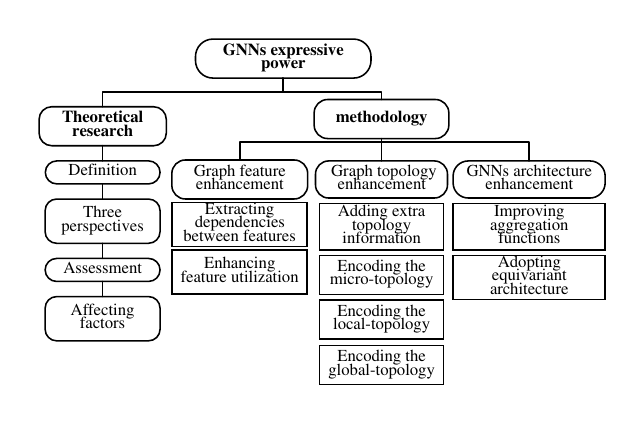}
\caption{The specific logical structure of the article.}
\label{fig_paperlogic}
\end{figure}

This work is organized as follows: Section \ref{sec2} lays out foundational knowledge on GNNs and graph isomorphism. Section \ref{sec3} delves into the importance of GNNs expressive power, offering a unified definition, research angles, evaluation metrics, and affecting factors. Section \ref{sec4} presents strategies to improve the expressive power, overviewing the high-expressivity GNNs with their pros and cons. Section \ref{sec5} discusses the challenges and opportunities in research. Finally, we conclude with Section \ref{sec6}, with the article's logical flow depicted in Figure \ref{fig_paperlogic}.

\section{Preliminary}\label{sec2}
Before presenting the relevant definitions and the problem description, Table \ref{table_1} gives a summary of all the notations used in the article. 

\begin{table}[!t]
\renewcommand{\arraystretch}{1.3}
\caption{Notations in the survey}
\label{table_1}
\centering
\begin{tabular}{|c||c|}
\hline
\textbf{Notations} & \textbf{Descriptions}\\
\hline
\{\} & A set.\\
\hline
\{\{\}\} & A multiset.\\
\hline
$G=(\mathcal{V},\mathcal{E})$ & An unfeatured graph.\\
\hline
$\mathcal{V}$ & The set of nodes in a graph. \\
\hline
$\mathcal{E}$ & The set of edges in a graph.\\
\hline
$N(v)$ & The set of the neighboring nodes of node $v$. \\
\hline
$G=({\bf A},{\bf X})$ & An featured graph.\\
\hline
${\bf A}$ & The adjacency matrix.\\
\hline
${\bf X}$ & The feature matrix. \\
\hline
${\bf h}_v^{(l)}$ & The embedding of node $v$ in the $l$-th layer.\\
\hline
${\bf H}^{(l)}$ & The ebbedding matrix in the $l$-th layer. \\
\hline
${\rm MSG}^{(l)}$ & The message passing function in the $l$-th layer.\\
\hline
${\rm AGG}^{(l)}$ & The aggregation function in the $l$-th layer.\\
\hline
${\rm UPD}^{(l)}$ & The update function in the $l$-th layer.\\
\hline
\end{tabular}
\end{table}

\subsection{Basics of Graph Neural Networks}
{\bf Graph.} In the following, we consider the problem of representation learning on one or several graphs of possibly varying sizes. Let $G=(\mathcal{V}, \mathcal{E}) \in \mathcal{G}$ be a graph of $n$ nodes with node set $\mathcal{V}$ and edge set $\mathcal{E}$, directed or undirected. Each graph has an adjacency matrix ${\bf A} \in \{0,1\}^{n \times n}$, and potential node features ${\bf X} = ({\bf x}_1, ..., {\bf x}_n)^{\mathrm{T}} \in {\mathbb{R}}^{n \times c}$. Combining the adjacency matrix and node features, we may also denote $G = ({\bf A}, {\bf X})$. When explicit node features are unavailable, an unfeatured graph $G$ will consider ${\bf X} = {\bf 1}{\bf 1}^{\mathrm{T}}$ where ${\bf 1}$ is a $n \times 1$ vector of ones. For a node $v \in \mathcal{V}$, the set of neighboring nodes is represented as $N(v)$. We also use $\mathcal{V}[G]$ to denote the node set of graph $G$.

\noindent {\bf Message Passing Neural Networks (MPNNs).} Out of the numerous architectures proposed, MPNNs have been widely adopted~\cite{gilmer2017neural,scarselli2008graph,battaglia2018relational}. MPNNs employ a neighborhood aggregation strategy to encode each node $v \in \mathcal{V}$ with a vector representation ${\bf h}_v$. This node representation is updated iteratively by gathering representations of its neighbors before performing a nonlinear transformation of them with neural network (NN) layers:
\begin{small}
\begin{equation}
{\rm Message Passing}: {\bf m}_{u}^{(l)}= {\rm MSG}^{(l)}\left(\left\{{\bf h}_{u}^{(l-1)}: u \in N(v)\right\}\right),
\end{equation}
\end{small}
\begin{small}
\begin{equation}
{\rm Aggregation}: {\bf a}_{v}^{(l)}= {\rm AGG}^{(l)}\left(\left\{{\bf m}_{u}^{(l-1)}: u \in N(v)\right\}\right),
\end{equation}
\end{small}
\begin{small}
\begin{equation}
{\rm Update}: {\bf h}_{v}^{(l)}={\rm UPD}^{(l)}\left({\bf h}_{v}^{(l-1)}, {\bf a}_{v}^{(l)}\right).
\end{equation}
\end{small}
where, ${\bf h}_{v}^{(0)}={\bf x}_{v}$, ${\bf m}_{u}^{(l)}$ represents the message from the neighbors, ${\bf m}_{v}^{(l)}$ represents the aggregated message from neighbors at layer $l$, ${\rm MSG}^{(l)}$, ${\rm AGG}^{(l)}$ and ${\rm UPD}^{(l)}$ denote the message passing, aggregation and update functions of layer $l$.

The neighbor aggregation strategy employed by MPNNs allows them to learn relationships between nearby nodes. This inductive bias is well suited for problems requiring relational reasoning~\cite{yoon2019inference,garcia2019combining,khalil2017learning,battaglia2016interaction,xu2019can}. Despite their widespread success, MPNNs nonetheless suffer from several limitations, including over-smoothing~\cite{chen2020measuring,oono2019graph,zhang2020revisiting}, over-compression~\cite{alon2020bottleneck,topping2021understanding}, the failure to distinguish between node identity and location~\cite{you2021identity,you2019position}, and limited expressive power~\cite{xu2018powerful,morris2019weisfeiler}. The lack of expressive power is evident in their inability to compute several crucial graph properties (e.g., shortest/longest cycles, diameters and local clustering coefficientss\cite{garg2020generalization}), and their failure to learn the graph topology (for example, whether a graph is connected or whether a particular pattern, such as a cycle, exists in the graph ~\cite{chen2020can}). To address tasks where the graph topology is critical, such as predicting the chemical properties of molecules\cite{elton2019deep,sun2020graph} and solving combinatorial optimization problems~\cite{mazyavkina2021reinforcement, vesselinova2020learning}, more powerful GNNs are required.

\noindent {\bf k-order Linear Invariant GNN (k-IGN) Framework.} The k-IGN by Maron is another widely used framework for investigating GNN's expressive power~\cite{maron2018invariant,maron2019universality}. The k-IGN comprises of alternating layers composed of invariant or equivariant linear layers and nonlinear activation layers. Each linear function is given as $L$, and the non-linear activation function is denoted as $\sigma$. The parameter k in k-IGN refers to the order of the tensor ${\bf A}$, which corresponds to the indices used to represent its elements. For instance, a first-order tensor represents the node feature, a second-order tensor represents the edge feature, and a $k$-th order tensor encodes hyper-edge feature. For $k=2$, the linear layer consists of 15 equivariant layers and 2 invariant layers, which can be expressed as follows:
\begin{small}
\begin{equation}
L({\bf A})=\sum_{i=1}^{15} \theta_{i} L_{i}({\bf A})+\sum_{i=16}^{17} \theta_{i} \bar{L}_{i}({\bf A})
\end{equation}
\end{small}

After adding the non-linear activation layer, the 2-IGN architecture can be presented as:
\begin{small}
\begin{equation}
{\bf A}^{(l+1)}=\sigma\left(L^{(l+1)}\left({\bf A}^{(l)}\right)\right)
\end{equation}
\end{small}

IGN has demonstrated not less expressive power than MPNN and is therefore being widely adopted to augment expressiveness. Aside from the k-IGN frameworks, there also have been efforts to construct more powerful equivariant networks by combining a set of simple permutation equivariant operators~\cite{barcelo2020expressive,barcelo2020logical,grohe2021logic,geerts2022expressive,balcilar2021breaking,geerts2022expressiveness}.

\noindent {\bf Permutation Invariance.} Graphs are inherently disordered, with no fixed node order. As a basic inductive bias in graph representation learning, permutation invariance refers to the fact that the output of a model is independent of the input order of nodes.  Specifically, a model $f$ is permutation invariant if it satisfies $f \circ \pi(G) = f(G)$ for all possible permutations $\pi$.

\noindent {\bf Permutation Equivariance.} On the other hand, permutation equivariance means that the output order of the model corresponds to the input order. When we focus on node features, we expect the model to output in the same order as the input nodes. Specifically, a model $g$ is permutation invariant equivariant if it satisfies $g \circ \pi(G) = \pi \circ g(G)$ for all possible permutations $\pi$.

The design philosophy of GNNs is to generate a permutation equivariant function on the graph by applying a local permutation invariant function to aggregate the neighbor features of each node~\cite{maron2018invariant}.

\subsection{Basics of Graph Isomorphism}
{\bf Graph Isomorphism.} Graph isomorphism refers to the topological equivalence of two graphs. Given two graphs $G=({\bf A}, {\bf X})$ and $G'=({\bf A'}, {\bf X'})$ with adjacency matrices ${\bf A}$ and ${\bf A'}$ and feature matrices ${\bf X}$ and ${\bf X'}$ respectively, an isomorphism between these graphs is a bijective mapping $\pi: \mathcal{V}[G] \longrightarrow \mathcal{V}[G']$ that satisfies ${\bf A}_{uv} = {\bf A'}_{\pi (u) \pi (v)}$ and ${\bf X}_{v} = {\bf X'}_{\pi (v)}$. We use $G \cong G'$ to denote that $G$ and $G'$ are isomorphic, where $G'$ can be represented as $\pi (G)$. If there is no bijection $\pi$ that satisfies these conditions, we say that $G$ and $G'$ are non-isomorphic.

\begin{figure}[!t]
\centering
\includegraphics[width=0.5\textwidth]{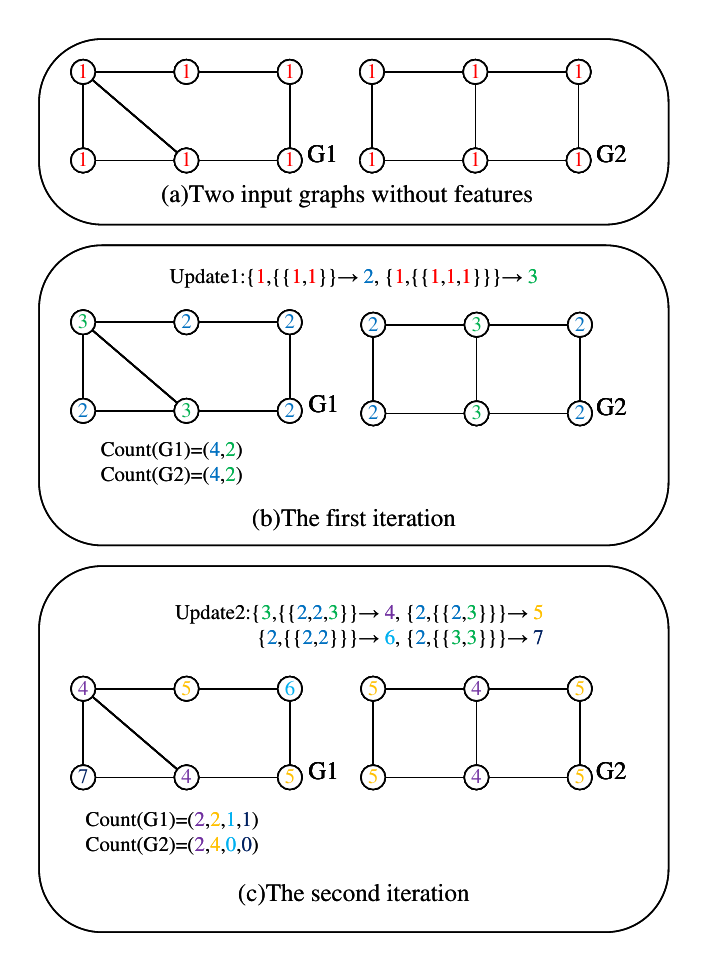}
\caption{The illustration of the aggregation and update process of WL test. (a) Given two graphs without features, and add color labels to all nodes. (b) In the first iteration, the different information aggregated by the nodes is mapped into new color labels, and then these new labels are redistributed to nodes, and the number of labels is counted after the distribution. After 1st iteration, $G1$ and $G2$ have the same color distributions to determine whether they are isomorphic, and the next iteration is performed. (c) Perform the node neighbor aggregation and redistribution of color labels steps again, and get the different color distributions of $G1$ and $G2$, at this time, it can be determined that the two are non-isomorphic.}
\label{wl}
\end{figure}

\noindent {\bf Weisfeiler-Lehman test (WL test).} Graph isomorphism problems are notoriously difficult and are classified as NP-hard in graph theory. The WL test of isomorphism~\cite{weisfeiler1968reduction,babai1979canonical}, also known as the color refinement algorithm~\cite{read1977graph}), is an effective algorithm to solve graph isomorphism problems. The WL test utilizes a graph coloring technique to assign color labels to each node in the graph. This iterative process is illustrated in Figure \ref{wl}\cite{shervashidze2011weisfeiler}. Initially, every node in the graph is assigned a unique color label ${\bf c}$. In each iteration, the central node and its neighbors are combined to form a set of multiple colors, which are subsequently mapped to a new unique color using Hash function, which becomes the new label of the central node. The process is repeated until the node color labels no longer change, resulting in a color distribution histogram of the graph. By comparing the color distributions of two graphs, the WL test can determine whether they are isomorphic or not. This process can be divided into several steps, neighbor label aggregation, multi-sets sorting, label compression, and label update and can be mathematically formulated as follows:
\begin{small}
\begin{equation}
{\bf c}_{v}^{(l)}={\rm Hash}\left({\bf c}_{v}^{(l-1)}, \left\{{\bf c}_{u}^{(l-1)}| u \in N(v)\right\}\right).
\end{equation}
\end{small}
 
It is important to note that identical coloring results could only suggest but cannot guarantee graph isomorphism.

While the WL test is fast and efficient for most graphs~\cite{babai1980random}, it has limitations in distinguishing members of many important graphs, such as regular graphs. The more powerful k-WL algorithm was initially introduced by Weisfeiler~\cite{weisfeiler2006construction}, expanding the coloring of individual vertices to the coloring of k-tuples. A k-tuple is a unit of k neighboured nodes, and the k-WL test~\cite{keriven2019universal, azizian2020expressive} captures higher-order interactions in the graph by considering the process of iterative aggregation between all neighboured k-tuple units. Specifically, the labels for tuple units consisting of k nodes are computed, treating these tuple units as significant graph entities for iterative operations. The k-WL becomes increasingly expressive as k grows. For sufficiently large k, it can distinguish any finite set of graphs. It has been observed that the discrimination potential for k$>$1, (k+1)-WL test is unequivocally stronger than the k-WL test, creating a hierarchical structure where the discriminative power increases with higher k values. Therefore, the k-WL test excels at detecting subgraph topology with larger k values, exhibiting an enhanced discrimination ability concerning graph structures.

K-WL iteratively recolors all $n^k$ k-tuples defined on a graph $G$ with $n$ nodes. In iteration 0, each k-tuple $\Vec{v}=(v_1,...,v_k) \in V(G)^k$ is initially assigned a color as its atomic type $c^{(0)}_k( \Vec{v})$. If $G$ has $d$ colors, then $c^{(0)}_k( \Vec{v})$ is an ordered vector encoding. Let $c^{(l)}_
k(\Vec{v})$ denote the color of k-tuple $\Vec{v}$ on graph $G$ at the t-th iteration of the k-WL. In the l-th iteration, k-WL updates the color of each $\Vec{v} \ in V(G)^k$ according to:
\begin{small}
\begin{equation}
\begin{split}
c_k^{(l+1)}(\Vec{v}) &= {\rm Hash}(c_k^{(l)}(\Vec{v}),\\
&\{\{c_k^{(l)}(\Vec{v}[x/1]|x \in V(G))\}\},...,\\
&\{\{c_k^{(l)}(\Vec{v}[x/k]|x \in V(G))\}\}).
\end{split}
\end{equation}
\end{small}

Let $c^{(l)}_k (G)$ denote the encoding of G at l-th iteration of k-WL. Then,
\begin{small}
\begin{equation}
c_k^{(l)}(G) = {\rm Hash}(\{\{c_k^{(l)}(\Vec{v})|\Vec{v} \in V(G)^k\}\}).
\end{equation}
\end{small}

Moreover, k-WL has interesting connections to logic, games, and linear equations~\cite{cai1992optimal, grohe2021logic, grohe2015pebble}. Another variation of k-WL is known as k-folklore WL (k-FWL), where it is shown that the expressive power of (k+1)-WL is equivalent to that of k-FWL\cite{shervashidze2011weisfeiler}. 

\section{The description of GNNs expressive power}\label{sec3}
\subsection{The necessity of GNNs expressive power}
\begin{figure*}[!t]
\centering
\includegraphics[scale=0.75]{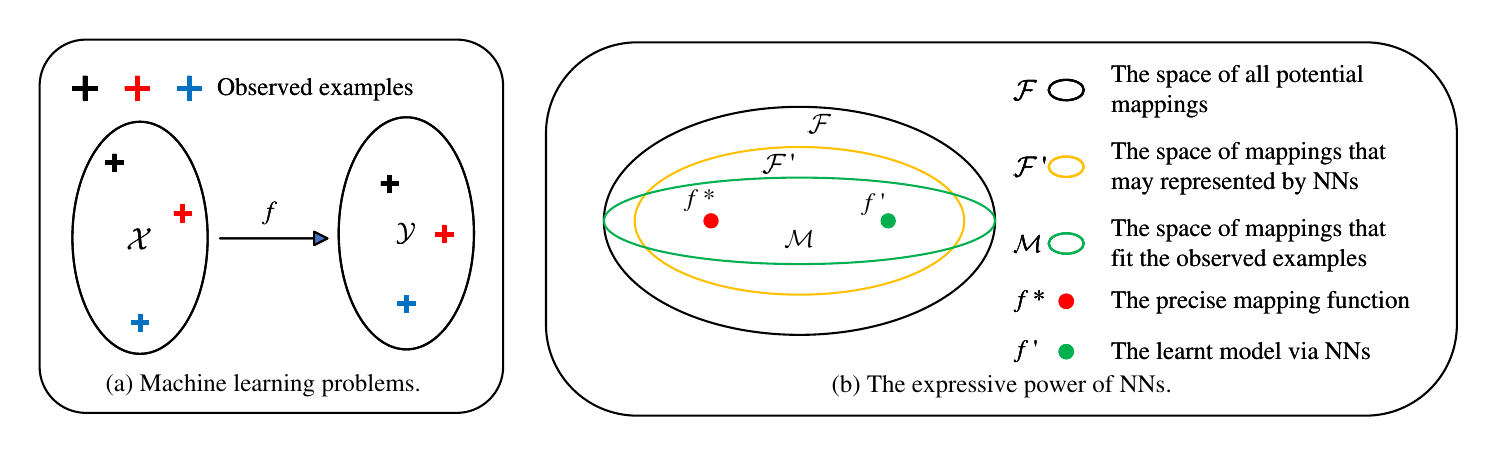}
\caption{An illustration of the expressive power of NNs and their effects on the performance of learned models. (a) Machine learning problems aim to learn the mapping from the feature space to the target space based on several observed examples. (b) The expressive power of NNs refers to the gap between the two spaces $\mathcal{F}$ and $\mathcal{F'}$. Although NNs are expressive ($\mathcal{F'}$ is dense in $\mathcal{F}$), the learned model $f'$ based on NNs may differ signifcantly from $f^{\ast}$ due to their overfit of the limited observed data. This figure is with reference to Figure. 5.5 in \cite{wu2022graph}.}
\label{fig_MLexpress}
\end{figure*}

{\bf Expressive power.} All machine learning problems can be abstracted as learning a mapping $f^{\ast}$ from the feature space $\mathcal{X}$ to the target space $\mathcal{Y}$. $f^{\ast}$ is usually approximated using a model $f_\theta$ by optimizing some parameters $\theta$. In practice, $f^{\ast}$ is typically a priori unknown, so one would like $f_\theta$ to approximate a wide range of $f^{\ast}$ as much as possible. An estimate of how wide this range is called the \textit{expressive power} of the model, providing an important measure of the model potential\cite{wu2022graph}, shown in Figure \ref{fig_MLexpress} (a).

The powerful expressive power of NNs is reflected in their ability to approximate all continuous functions~\cite{cybenko1989approximation}, specifically the ability to embed data within the feature space $\mathcal{X}$ into the target space $\mathcal{Y}$ generated by any continuous function, which is in fact feature embedding ability, show in Figure \ref{fig_MLexpress} (b). Due to the strong expressive power of NNs, similarly, few works~\cite{zhang2018end, ying2018hierarchical, verma2018graph, santoro2018measuring} doubt the expressive power of GNNs that have demonstrated significantly superior performance in a variety of application tasks because they naturally feature the superior performance of GNNs to their excellent feature embedding ability.

However, some graph-augmented multilayer perceptron (MLP) models~\cite{gasteiger2018predict, frasca2020sign} outperform GNNs despite the former's less expressive power than the latter in a number of node classification problems~\cite{chen2020graph}, where MLPs~\cite{rumelhart1986learning} use only information from each node to compute feature embeddings of nodes, and GNNs iteratively aggregate the features of neighbouring nodes in each layer, thus allowing the use of global information to compute feature embeddings of nodes. This fact results in a contradiction between our intuitive understanding of expressive power, yet at the same time illustrates that feature embedding power does not describe the expressive power of GNNs well.

GNNs, compared to NNs, add the inductive bias of permutation invariance so that they can propagate and aggregate information over the topology of the graph. ~\cite{oono2019graph, yang2022graph} demonstrate that if GNNs only have feature embedding ability but lack the ability to maintain the graph topology, then their performance may be inferior to MLP. In addition, we know that the expressive power of feed-forward neural networks is often limited by their width, whereas the GNNs' expressive power is not only limited by their width, but also how they use the graph topology for message propagation and node updates. It follows that the topology representation ability of GNNs, i.e., the ability to maintain the graph topology, becomes the key to their superior performance. To characterise the expressive power of GNNs using their topology representation ability then requires a new set of theoretical tools.

\subsection{The definition of GNNs expressive power}
Graph data typically contains both feature and topology information. Therefore, the expressive power of GNNs refers to the ability of the model to encode and process feature information on the graph as well as the structural information of the graph. A GNN model with high expressive power can not only capture the complex relationships in the feature, but also the patterns of the graph structure, and then transform them into meaningful representations. The level of expressive power directly affects the performance and effectiveness of the model. Here, the feature embedding ability represents the part of GNNs model that encodes and processes feature information, and the topology representation ability embodies the model's ability to encode and process structural information. As a neural network model for graph representation learning in the family of NNs, the feature embedding ability of GNNs is unquestionable, and their expressive power is more reflected in their topology representation ability. GNNs map node $v$ from original space $\mathcal{X}$ to embedding space $\mathcal{Y}$ to obtain node embedding ${\bf h}_v$ via function $f$, where ${\bf h}_v = f({\bf A} \cdot {\bf x}_v)$. For nodes $v_1$, $v_2$ in different topological positions on graph $G$, if ${\bf x}_1 \neq {\bf x}_2$, the results of the feature embedding ability ensure that $f({\bf A} \cdot {\bf x}_1) \neq f({\bf A} \cdot {\bf x}_2)$. And if ${\bf x}_1 = {\bf x}_2$, to maintain the topology of the graph, it meanwhile requires that $f({\bf A} \cdot {\bf x}_1) \neq f({\bf A} \cdot {\bf x}_2)$. The relevant formal definitions are further given below.

{\bf Feature embedding ability.} Let ${\bf X}={\bf random}$, i.e., input graph with node features, the feature embedding ability can be expressed by the size of the value domain space $\mathcal{F} = \{f({\bf A} \cdot {\bf X})|{\bf X}={\bf random}\}$ of $f$.

{\bf Topology representation ability.} Let ${\bf X}={\bf 1}$, , i.e., input graph without node features, the topology representation ability can be expressed by the size of the value domain space $\mathcal{F'}= \{f({\bf A} \cdot {\bf X})|{\bf X}={\bf 1}\}$ of $f$.

\begin{figure*}[!t]
\centering
\includegraphics[scale=0.75]{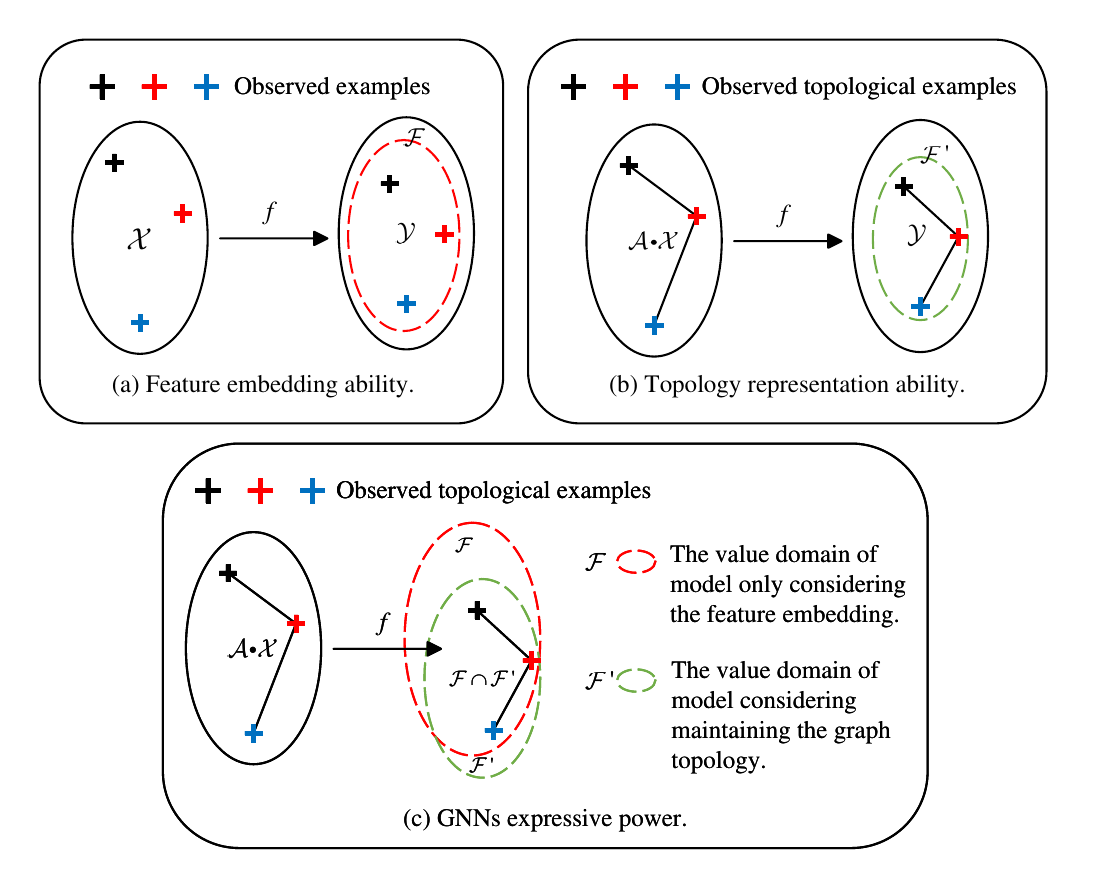}
\caption{An illustration of the expressive power of GNNs. a) The feature embedding ability of GNNs is the same as that of NNs in that both maps the observed examples in the feature space $\mathcal{X}$ to the target space $Y$ via $f$. The strength of the feature embedding ability is measured by the size of the value domain space $\mathcal{F}$ of $f$. b) The topology representation capability of GNNs is achieved by mapping the observed examples in the feature space to the target space via $f$ and preserving the original topology between the examples. The strength of the capability is measured by the size of the value domain space $\mathcal{F'}$ with ${\bf X}={\bf 1}$. c) The expressive power of GNNs consists of a combination of feature embedding ability and topology representation ability, measured by the size of the intersection of $\mathcal{F}$ and $\mathcal{F'}$ with ${\bf X}={\bf random}$.}
\label{fig_def}
\end{figure*}

The feature embedding ability reflects that nodes with different features can get different node embeddings. Topology representation ability reflects that nodes with different topological positions can get different node embeddings. Together, these two constitute the expressive power of GNNs, illustrated with reference to Figure \ref{fig_def}.

{\bf The expressive power of GNNs.} Let ${\bf X}={\bf random}$, the size of the space $\mathcal{F} \cap \mathcal{F'}$ describes the expressive power of $f$. Typically, $\mathcal{F} \supseteq \mathcal{F'}$, which means that the topology representation ability is the main factor limiting the expression of GNNs. And for two models, $f_1$ and $f_2$, if $\mathcal{F}_1 \supset \mathcal{F}_2$, we say that $f_1$ is more expressive than $f_2$.

\subsection{The existing perspectives of GNNs expressive power}
Several prevailing views in current research on the expressive power of GNNs characterise the expressive power as approximation ability, separation ability and subgraph counting ability, respectively.

{\it Approximation ability}. This perspective follows the path of classical neural network expressive power research, which uses the ability of GNNs to approximate functions on graphs to characterize expressive power, and the focus of the research is actually on examining the feature embedding ability of the model. In traffic network flow prediction, GNNs update the representation of each road node by aggregating the flow information of neighboring road sections and considering time series features. In this way, GNNs learn a function capable of mapping current and past traffic states to future flows. A GNN model with high approximation capability is able to accurately predict future flows because it is able to effectively understand and model complex traffic flow patterns and interactions between road networks. The higher approximation ability ensures that GNNs are able to process graph data more efficiently while maintaining a high level of accuracy. However, further exploration of the types of functions that GNNs can approximate is still insufficient to solve more and more complex practical problems.

{\it Separation ability}. This perspective builds on the framework of GNNs expressive power research based on the WL test, which uses the ability of GNNs to recognize non-isomorphic graphs to describe expressive power, and the focus of the research is actually on examining the topology representation ability of the model. This ability is important when dealing with large graphs with complex structures. For example, in social networks, people may have multiple different social circles, and GNNs need to be able to recognize these circles and separate them. However, the scope of research on specific separation ability is limited to featureless graphs, which leads to their lack of practical applicability in solving real-world problems.

{\it Subgraph counting ability}. This perspective also lies in examining the topology representation ability of models, using the ability of GNNs to recognize and count substructures on graphs to describe the expressive power. Unlike separation ability, which focuses on structural differences between pairs of graphs, subgraph counting ability focuses more on distinguishing subgraph structures within a single graph. Unlike the whole graph structures, subgraph structures are highly relevant to molecular functional groups and communities in real-world tasks, such as predicting the properties of a molecule in chemistry, which requires counting different structures in a molecule\cite{gilmer2017neural}, and thus have a higher value for practical applications. In predicting specific properties of a molecule, GNNs identify specific substructures in the molecule, such as benzene rings or hydroxyl groups, by learning the topology of the graph. Then, by counting the number of these substructures in the molecule, GNNs can capture key information that is closely related to the properties of the molecule. For example, the number of benzene rings may affect the stability and reactivity of the molecule; the presence and number of hydroxyl groups may determine the solubility of the molecule.

\subsubsection{Approximation ability}
Models with stronger expressive power can learn more complex mapping functions. Neural networks are known for their strong expressive power, and one of the most famous results is the universal approximation theorem for neural networks\cite{cybenko1989approximation, cybenko1992approximation, hornik1991approximation}. As a member of the neural network family, GNNs can also use the ability to approximate mapping functions to describe their own expressive power. However, because GNNs add inductive bias\cite{battaglia2018relational} to graph learning, i.e., permutation invariance, the results of the generalized approximation theorem for neural networks cannot be simply generalized to GNNs. In this context, Maron et al.~\cite{maron2019universality, azizian2020expressive, keriven2019universal} describe specifically the types of functions that GNNs can approximate, using the {\it approximate ability} to approximate to describe the expressive power of GNNs, that is, what kind of permutation invariant graph functions GNNs can approximate. They further give a description of the set of functions that the model can approximate as a measure of expressive power. Assume that GNN is a mapping  $f_\theta: \mathcal{G} \rightarrow R^{d \times n \times m}$ on graph, where $\mathcal{G}$ is the set of graphs $G=({\bf A}, {\bf X})$, ${\bf X} \in R^{d \times n}$ is node feature, $d$ is the dimension of the feature, $n=|V|$ and $m=|\mathcal{G}|$. Denote the output of GNNs by $f_\theta (G)$, the formal definition of approximate ability can be given.

{\bf Function approximation.} Consider a function $f^{\ast}$ to be learned, when the input of GNNs is the set of graphs $\mathcal{G}$, if $||f^{\ast}(G) - f_\theta (G)||_\infty := sup_G |f^{\ast}(G) - f_\theta (G)| < \varepsilon$, for $G \in \mathcal{G}$ and $\varepsilon >0$, then $f^{\ast}$ is successfully approximated by $f_\theta$.

{\bf Approximation ability.} The measure of the size of the set of functions composed of all potential $f_\theta$ is the approximation ability of GNNs. Figure \ref{fig_def_inout} shows the representation of input and output under the definition of approximation ability.

However, these theoretical findings are not sufficient to explain the unprecedented success of GNNs in practice, because simply knowing what graph functions the model can approximate does not guide practical production work. It is challenging to analyze the expressive power of GNNs starting from a practical application perspective.

\subsubsection{Separation ability}
Considering how GNNs make predictions about molecules, GNNs should be able to predict whether a topology corresponds to a valid molecule by identifying whether the topology of the graph is the same or similar to the topology corresponding to a consistent valid molecular. This problem of measuring whether two graphs have the same topology involves the problem of graph isomorphism ~\cite{helfgott2017graph, levin2000stochastic}.

In the graph isomorphism perspective, Xu et al.\cite{xu2018powerful} and Morris et al.\cite{morris2019weisfeiler} used {\it separation ability} to describe the expressive power of GNNs, that is, whether GNNs can effectively identify non-isomorphic graph topology. By gaining insight into the high similarity and close connection between the WL test and GNNs in the iterative aggregation process, they further used the WL test as a measure of the strength of the separation ability and indicated that the upper limit of the separation ability is 1-WL test. Similarly, the formal definition of the separation ability can be given.

{\bf Graph isomorphism recognition.} Consider two non-isomorphic graphs $G_1 = ({\bf A}_1, {\bf X}_1)$ and $G_2=({\bf A}_2, {\bf X}_2)$, where $X_1=X_2={\bf 1}$. If $f_\theta (G_1) \neq f_\theta (G_2)$, then it is assumed that GNNs can identify non-isomorphic graph pairs.

{\bf Separation ability.} The measure of the size of the set of all identifiable possible non-isomorphic pairs $|\mathcal{G}_1|$ is the separation ability of GNNs. Figure \ref{fig_def_inout} shows the representation of input and output under the definition of separation ability.

The results of the separation ability show the role that GNNs expressive power plays in practice. And in fact, it has been shown that both separation and approximation ability are theoretically unified. Chen et al.\cite{chen2019equivalence} demonstrated the equivalence between the two abilities, showing that the basis of the approximation ability is actually the embedding of two non-isomorphic graphs into different points of the target space, and that a model that can distinguish all nonidentical graphs is also a general function approximator. The result of equivalence provides, in a sense, a practical application for approximation ability.

However, the result of separation ability requires comparison of pairs of graphs, which requires a prior a different GNN for each pair of graphs to distinguish them. This is of little help in practical learning scenarios, because people prefer to give meaningful representations to all graphs through a single GNN. In addition, the graph isomorphism problem measures the expressive power at the graph level, while the expressive power at the node level is equally important, and whether this theory can explain the expressive power of nodes is still questionable\cite{Benjamin2021}.

\subsubsection{Subgraph counting ability}
Considering solving problems at the node level is also very instructive for practice. Continuing with the prediction problem of predicting molecular functions, in addition to the need to identify whether the graph topology is the same as the known functional group topology to determine the molecular properties, sometimes the number of functional groups also has an impact on the function of the molecule, which requires determining how many identical graph topology there are in the graph. To address this issue, Chen et al.\cite{chen2020can} have proposed to use {\it subgraph counting ability} to describe the GNNs expressive power, emphasizing the GNNs' ability to recognize subgraph topology in graphs. On the one hand, this ability can evaluate node-level embeddings, comparable to the separation ability. On the other hand, real-world graph topology such as molecular and community topology typically involve subgraph topology, Hence this ability has more practical guidance for applications compared to the approximation ability. Similarly, the formal definition of the separation ability can be given.

{\bf Subgraph counting.} Consider two isomorphic subgraphs $G_1 = (V_1, E_1)$ and $G_2 = (V_2, E_2)$ of graph $G = (V, E)$, where $V_1, V_2 \in V$, $E_1, E_2 \in E$ and $V_1 \neq V_2$. If $f_\theta (G_1) \neq f_\theta (G_2)$, then it is assumed that GNNs can count subgraphs.

\begin{figure*}[!t]
\centering
\includegraphics[scale=0.75]{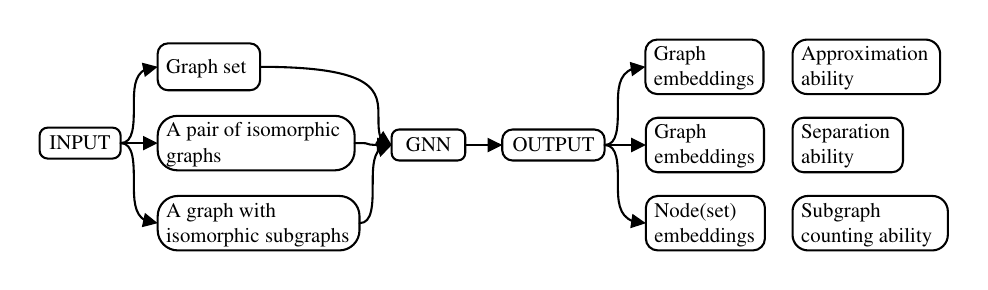}
\caption{Input and output of the GNNs model under different expressive power representations. When approximate ability is used to characterise the expressive power, the input of the model is a set of graphs and the output is a graph embeddings. When separation ability, the input is a pair of graphs and the output is graph embeddings. When the subgraph counting ability, the input is a single graph and the output is a node (set) embeddings.}
\label{fig_def_inout}
\end{figure*}

{\bf Subgraph counting ability.} The measure of the size of the set of all identifiable countable isomorphic subgraphs $|\mathcal{G}_1|$ is the subgraph counting ability of GNNs. Figure \ref{fig_def_inout} shows the representation of input and output under the definition of subgraph counting ability.

\subsection{The assessment of GNNs expressive power}
The current assessment of the strength of the expressive power of GNNs focuses more on assessing the strength of the topology representation of the model. According to the results of separation ability, the upper limit of the expression power of GNNs is the 1-WL test, the conclusion that comes from the observations of Xu et al.\cite{xu2018powerful} and Morris et al.\cite{morris2019weisfeiler}, who were the first to gain insights into the high degree of similarity between the neighborhood aggregation operation of GNNs and the aggregation of neighborhood labels in the WL test, and to use the WL test to assess the strength of the expressive power. However, as more and more models beyond the upper limit of the 1-WL test's ability were developed, the results of higher-order WL tests were used to assess the expressive power of these models, and the upper limit of expressive power was extended to the 3-WL test. However, the results of the k-WL test can only provide a simple general classification of the model's expressive power, and descriptive evaluation results such as "equivalent to the 1-WL test" and "not exceeding the 3-WL test" are useless in guiding the development and design of models and their practical applications scientifically and accurately. To address this issue, Puny et al.\cite{puny2023equivariant} introduced polynomial expressivity, utilizing graph polynomials to measure the expressive power of GNNs. This method quantifies both node and edge-level expressive power by considering polynomial functions on the adjacency matrix. Although it offers a detailed evaluation, its reliance on complex mathematics may limit its capability in fully capturing all graph structural features. Building on this concept, Zhang et al.\cite{zhang2024beyond} proposed homomorphic expressivity. This technique evaluates a model's capacity to encode substructures under homomorphic counting, providing a comprehensive assessment tool for comparing the expressive power of different GNN models directly.

While the WL hierarchy, polynomial expressivity, and homomorphic expressivity offer insights into assessing the expressive power, they primarily concentrate on graph structure, neglecting the importance of feature information. Recognizing that topological representation alone is insufficient, the role of feature embedding in enhancing model performance is also crucial. Interaction between features can effectively improve model performance~\cite{9944200}.  Furthermore, GNNs topology representation ability is also significantly improved after considering into the feature embedding ability. A compelling demonstration by Kanasoulis et al.\cite{kanatsoulis2022graph} showed the ability of gnn to recognize any graph with at least one node displaying unique features. This finding reaffirms our reasonable definition of the expressive power of GNNs, emphasizing that feature embedding supports the topology representation of GNNs. In particular, if the node features adequately describe the node constants, Loukas et al.~\cite{loukas2019graph, vignac2020building, bodnar2022neural} confirmed that GNNs with strong message passing and update functions become Turing universal. That is, if the feature information is sufficient for us to distinguish between nodes, the structural information plays little role in the process of node differentiation. This is confirmed in a small example of the graph isomorphism problem\cite{loukas2020hard}.

Strong topology representation ability and strong feature embedding ability determine the power of GNNs models to be efficiently expressed, allowing them to perform well in downstream tasks.  Currently, the research community has not yet developed an evaluation method that can simultaneously and comprehensively assess the ability of GNNs to express both graph structure and feature information, and the comprehensive expressive power of GNNs is usually only indirectly measured through experimental evaluation.

\subsection{The factors affecting GNNs expressive power}
The expressive power of GNNs is influenced by both the chosen graph information encoding method and the design of the model aggregation function. The selection of the graph information encoding method, being a more overarching factor, impacts the representation of all GNN models. Conversely, the design of the model aggregation function is more focused and tailored, affecting the expressiveness of specific models within the GNN framework.

The way of encoding graph information affects expressive power by influencing the representation of graph features and graph topology. First, node feature information has strong discriminative power. GNNs can distinguish between graphs with different node features\cite{loukas2019graph} and any graph that differs in at least one node feature\cite{kanatsoulis2022graph}. Second, information about relationships between features also has recognition ability, and GNNs can distinguish graphs with different global graph features\cite{barcelo2020expressive}. In addition, structural information is also recognizable, and GNNs can distinguish a pair of non-isomorphic graphs\cite{xu2019can}, as well as recognize different subgraph structures on a single graph\cite{chen2020can}. The classical GNNs model only considers the coded node feature information, and the expressive power of the model can be further improved by further considering the relationship information between coded features and the graph structure information.

\begin{figure}[!t]
\centering
\includegraphics[scale=0.75]{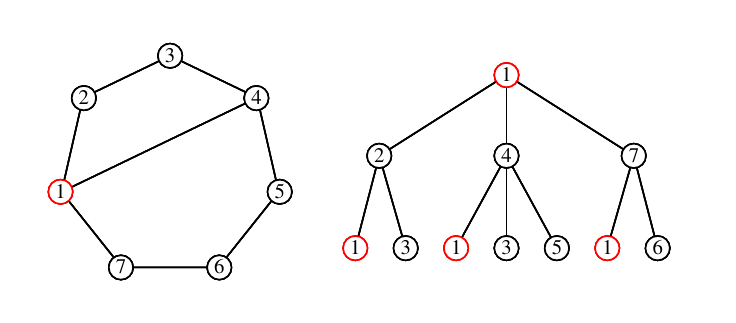}
\caption{The rooted subtree in message passing. The left one represents the original graph $G$, and the objective is to obtain an embedding of node $v_1$ on $G$, where node $v_1$ is denoted by red color. The right one is the subtree obtained for node $v_1$ after two rounds of iteration.}
\label{fig_subtree}
\end{figure}

The aggregation function also affects the representation of graph features and graph topology. First, GNNs do not directly encode the graph topology for learning, but use the topology of the graph for information transmission and aggregation, which is realized by the permutation invariant aggregation function. This design hinders the expression of GNNs\cite{de2020natural}. Because the permutation invariant aggregation function assumes that the states of all adjacent nodes are equal, the relationship between adjacent nodes is ignored. The node encoding obtained through neighborhood aggregation represents the rooted subtree around the central node, as illustrated in the Figure \ref{fig_subtree}. However, rooted subtrees have limited expressive power for representing non-tree graphs. Therefore, the central node cannot distinguish whether two adjacent nodes are adjacent or not, and cannot identify and reconstruct the graph topology. General GNNs can only explicitly reconstruct a star graph from a 1-hop neighborhood, but cannot model any connections between neighborhoods\cite{chen2020can}. Therefore, the permutation invariant aggregation functions cause GNNs to lose the topology of the context in the topology representation and thus fail to learn the basic topological properties of the graph\cite{garg2020generalization}. Secondly, under the condition that the permutation invariance of the aggregation function is guaranteed, the specific aggregation affects the model's handling of the node features, e.g., the maximum and minimum aggregation functions lead to a lower utilization of the node feature information in the model, while the mean value as well as other aggregation functions such as the attention mechanism increase the utilization of the node feature information and thus improve the model's representation. The classical GNNs model needs to maintain the permutation invariance of the aggregation function, based on which designing more efficient aggregation functions that utilize the feature information can improve the expressive ability of the model. In addition, considering other more effective means to maintain the permutation invariance of the graph instead of using the aggregation function can also further improve the expressive ability of the model.

\section{Existing work of improving GNNs expressive power}\label{sec4}
\subsection{Methods to improve GNNs expressive power}
Improving GNNs expressive power involves bolstering feature embedding and topology representation. Feature embedding enhancement involves extracting feature dependencies and adding related features. Topology representation can be improved by encoding topology directly or optimizing GNNs architecture to overcome limitations from permutation invariant aggregation functions. The strategies can be categorized into three methods:

{\bf Graph feature enhancement} aims to enrich feature embeddings by {\bf extracting dependencies between features} to obtain more informative and expressive embedding results, or by {\bf enhancing feature utilization} to obtain a comprehensive feature representation.

{\bf Graph topology enhancement} aims to encode graph topology by {\bf adding extra topology information} manually to improve nodes’ distinctiveness or directly encoding the graph topological structure. The latter can be classified into {\bf encoding micro-topology} (capturing distance or position information between nodes to enhance the perception of graph structure), {\bf encoding local-topology} (using subgraphs or higher-order connectivity patterns to identify complex topological patterns), and {\bf encoding global-topology} (directly capturing global structural attributes such as connectivity and clustering structure).

{\bf GNNs architecture enhancement} aims to address limitations in learning graph topology due to permutation invariant aggregation functions. This can be achieved by {\bf improving aggregation functions} to better capture topological information or by {\bf adopting equivariant architectures} to overcome permutation invariance limitations, enabling the computation of global features.

Consistent with expectations, the three aforementioned methodologies encompass all currently known model designs capable of improving expressive power. Table \ref{table_2} examines and systematically categorizes the design of more powerful expressive GNNs in recent years based on their employed design methodologies.

\begin{table}[!t]
\renewcommand{\arraystretch}{1.3}
\caption{Summary of the powerful model}
\label{table_2}
\centering
\resizebox{\linewidth}{!}{ 
\begin{tabular}{|c||c||c|}

\hline
\textbf{Methods} & \textbf{Subclasses} & \textbf{Models} \\
\hline
\multirow{5}{*}{\makecell[l]{Graph\\feature\\enhancement}} & \multirow{3}{*}{\makecell[l]{Extracting dependencies\\between features}} & AM-GCN\cite{wang2020gcn} \\
 &  & CL-GNN\cite{hang2021collective} \\
 &  & ACR-GNN\cite{barcelo2020expressive} \\
\cline{2-3}
 & \multirow{2}{*}{\makecell[l]{Enhancing\\feature utilization}} & \multirow{2}{*}{MM-GNN\cite{Bi2023MMGNN:MG}} \\
 &  & \\

\hline
\multirow{22}{*}{\makecell[l]{Graph\\topology\\enhancement}} & \multirow{5}{*}{\makecell[l]{Adding extra\\topology information}} & Twin-GNN\cite{wang2022twin} \\
 &  & ID-GNN\cite{you2021identity} \\
 &  & CLIP\cite{dasoulas2019coloring} \\
 &  & RP-GNN\cite{murphy2019relational} \\
 &  & RNI-GNN\cite{abboud2020surprising} \\
 \cline{2-3}
 & \multirow{5}{*}{\makecell[l]{Encoding micro-topology}} & DE-GNN\cite{li2020distance} \\
 &  & P-GNN\cite{you2019position} \\
 &  & PEG\cite{wang2022equivariant} \\
 &  & SMP\cite{vignac2020building} \\
 &  & GD-WL\cite{zhang2023rethinking} \\
\cline{2-3}
 & \multirow{2}{*}{\makecell[l]{Encoding global-topology}} & Eigen-GNN\cite{zhang2021eigen} \\
 &  & SAGNN\cite{zeng2023substructure} \\
\cline{2-3}
 & \multirow{12}{*}{\makecell[l]{Encoding local-topology}} & GSN\cite{bouritsas2022improving} \\
 &  & GraphSNN\cite{wijesinghe2021new} \\
 &  & GNN-AK\cite{zhao2021stars} \\
 &  & NGNN\cite{zhang2021nested} \\
 &  & MPSN\cite{bodnar2021weisfeiler} \\
 &  & ESAN\cite{bevilacqua2021equivariant} \\
 &  & LRP-GNN\cite{chen2020can} \\
 &  & k-GNN\cite{morris2019weisfeiler} \\
 &  & K-hop GNN\cite{nikolentzos2020k} \\
 &  & KP-GNN\cite{feng2022powerful} \\
 &  & N2-FWL\cite{feng2024extending} \\
 &  & PIN\cite{truong2024weisfeiler} \\

\hline
\multirow{13}{*}{\makecell[l]{GNNs\\architecture\\enhancement}} & \multirow{6}{*}{\makecell[l]{Improving\\aggregation function}} & GIN\cite{xu2018powerful} \\
 &  & modular-GCN\cite{dehmamy2019understanding} \\
 &  & diagonal-GNN\cite{kanatsoulis2022graph} \\
 &  & GNN-LF/HF\cite{zhu2021interpreting} \\
 &  & Geom-GCN\cite{pei2020geom} \\
 &  & PG-GNN\cite{huang2022going} \\
\cline{2-3}
 & \multirow{7}{*}{\makecell[l]{Adopting\\equivariant architecture}} & k-IGN\cite{maron2018invariant} \\
 &  & PPGN\cite{maron2019provably} \\
 &  & k-FGNN\cite{geerts2020expressive} \\
 &  & Ring-GNN\cite{chen2019equivalence} \\
 &  & SUN\cite{frasca2022understanding} \\
 &  & GNNML\cite{balcilar2021breaking} \\
 &  & k-MPNN\cite{geerts2022expressiveness} \\
\hline
\end{tabular}
}
\end{table}

\subsection{Graph feature enhancement}
\subsubsection{Extracting dependencies between features}
{\bf AM-GCN} The adaptive multi-channel graph convolution network (AM-GCN)\cite{wang2020gcn} computes node similarity based on their features, constructing a k-nearest neighbor(k-NN) graph by selecting the k most analogous neighbors for each node. Then subsequently performs convolutions over both the k-NN graph and original graph. This design captures the dependencies between node features, thus achieving an enhancement of node features. For a k-NN graph $G_f=({\bf A}_f,{\bf X})$, ${\bf A}_f$ denotes the adjacency matrix. And the message passing along $G_f$ can be represented as equation \ref{equa13}.
\begin{small}
\begin{equation}
{\bf h}_{f}^{(l+1)}={\rm RELU}\left(\tilde{{\bf D}}_{f}^{-1/2} \tilde{{\bf A}}_{f} \tilde{{\bf D}}_{f}^{-1 / 2} {\bf h}_{f}^{(l)} {\bf W}_{f}^{(l)}\right),
\label{equa13}
\end{equation}
\end{small}
where ${\bf W}_f$ is the weight matrix, $\tilde{{\bf A}}_{f}={\bf A}_{f}+{\bf I}_{f}$, and $\tilde{{\bf D}}_f$ is the diagonal degree matrix of $\tilde{{\bf A}}_f$.

{\bf CL-GNN} GNNs can extract more information regarding the local neighborhood dependencies of the nodes by sampling\cite{murphy2019relational}, while sampling the predicted node labels also allows GNNs to focus on the relationships between node features, graph topology, and label dependencies. \cite{hang2021collective} consider using a collective learning framework (CL-GNN), which employs Monte Carlo sampling to further consider label dependencies. This approach enables GNNs to incorporate more information into the estimated joint label distribution, leading to more expressive node embeddings.

{\bf ACR-GNN} Barceló et al.\cite{barcelo2020expressive} adds a readout function in the aggregation operation to calculate the global node features for node update, resulting in a novel architecture known as aggregate-combine-readout GNN (ACR-GNN). The update process can be described as follows:
\begin{small}
\begin{equation}
	\begin{split}
		{\bf h}_{v}^{(l+1)}= {\rm UPD}^{(l)}\left({\bf h}_{v}^{(l)}, {\rm AGG}^{(l)}\left(\left\{{\bf h}_{u}^{(l)}: u \in N(v)\right\}\right),\right.\\
		\phantom{=\;\;}
		\left. {\rm READ}^{(l)}\left(\left\{{\bf h}_{u}^{(l)}: u \in V\right\}\right)\right).
	\end{split}
\end{equation}
\end{small}

\subsubsection{Enhancing feature utilization}
{\bf MM-GNN} Bi et al.\cite{Bi2023MMGNN:MG} argued that most of the existing GNNs aggregating the features of the neighbors by a single statistic lose the information related to the feature distribution of the neighbors, and thus degrade the model performance. Inspired by the method of moments in statistical theory, they proposed to use multi-order moments to describe the distribution of neighborhood features, and based on this, they designed a new GNN model, i.e., a hybrid moment graph neural network (MM-GNN), by calculating the multiorder moments of each node's neighbors as a signature, and then assigning a larger weight to each node's significant moments and updating the node representations using an elemental-attention-based moment adapter. The update process can be described as follows:
\begin{small}
\begin{equation}
a_k^{(l+1)}\left(v_i\right)=\sigma\left(\left[h_i^{(l)} \cdot W_{\text {query }}^{(l+1)} \| M_k^{(l+1)}\left(v_i\right) \cdot W_{k e y}^{(l+1)}\right] \cdot W_a^{(l+1)}\right),
\end{equation}
\end{small}
\begin{small}
\begin{equation}
h_i^{(l+1)}=\sum_{k=1}^K a_k^{(l+1)}\left(v_i\right) \odot M_k^{(l+1)}\left(v_i\right), a_k^{(l+1)}\left(v_i\right) \in \mathbb{R}^{1 \times D_{\text {hidden}}},
\end{equation}
\end{small}
where $W_a^{(l)}$ is the learnable attention matrix in $l$-th layer, $W_{\text {query }}^{(l)}, W_{k e y}^{(l)}$ are the transformation matrix for the query and key vectors in $l$-th layer. $M_k^{(l+1)}\left(v_i\right)$ is the form of $k$-th order moment for $(l+1)$-th layer of GNNs. The $\odot$ is the element-wise product and $K$ is the hyper-parameter for the largest ordinal of moments used in MM-GNN.

\subsection{Graph topology enhancement}
\subsubsection{Adding extra topology information}
The aim of adding extra topology information is to increase the discrimination between nodes topologically. A prevalent and straightforward approach entails assigning distinct, recognizable features to nodes, such as identifications\cite{wang2022twin}, \cite{you2021identity}, colors\cite{dasoulas2019coloring}, and random features~\cite{murphy2019relational, loukas2019graph, sato2019approximation, sato2021random, abboud2020surprising}. The added features can be represented by a characteristic matrix ${\bf C}$, as demonstrated in the equation \ref{equa44}, which enhances the node features ${\bf X}$ via ${\bf C}$. Especially, this method can also be applied to unfeatured graphs, in which case the additional features serve as node features.
\begin{small}
\begin{equation}
G=({\bf A}, {\bf X} \oplus {\bf C}).
\label{equa44}
\end{equation}
\end{small}

{\bf Twin-GNN} Wang et al.\cite{wang2022twin} have introduced the Twin-WL test as an extension of the 1-WL test, with improved discriminative power obtained by simultaneously assigning color labels and identity $id$ to the nodes and executing two sets of aggregation schemes. The identity aggregation as Equation \ref{equa_idagg}. Based on the Twin-WL test, the Twin-GNN has been developed to operate two sets of message passing mechanisms for node features and node identities. The identity passing encodes more complete topology information, allowing for a stronger expression than 1-WL test without significantly increasing computational costs.
\begin{small}
\begin{equation}
{\bf id}_{v}^{(l)}=f_{id}\left({\bf id}_{v}^{(l-1)}, \left\{{\bf id}_{u}^{(l-1)}| u \in N(v)\right\}\right),
\label{equa_idagg}
\end{equation}
\end{small}
where $f$ is a function able to convert the identity multiset to the topology information that can be compared across different graphs.

{\bf ID-GNN} The Identity-aware GNNs (ID-GNNs)\cite{you2021identity} are a universal extension of GNNs. A L-layer ID-GNN first extracts the L-hop network $G^{(L)}_{v}$ of node $v$, and injects a unique color identity to $v$. Then, the heterogeneous message passing is performed on the $G^{(L)}_{v}$ to obtain the node embedding with injected identity. This added identification information enables the distinguishing of subgraphs with the same topology, thereby enhancing the GNNs expressive power.

For $u \in \mathcal{V}(G^{(L)}_{v})$, there can be categorized into two types throughout the embedding process: nodes with coloring and nodes without coloring. When passing the message of the colored nodes, perform another set of message passing mechanism:
\begin{small}
\begin{equation}
{\bf m}_{s}^{(l)}= {\rm MSG}^{(l)}_{\rm color}\left(\left\{{\bf h}_{s}^{(l-1)}: s \in N(u)\right\}\right).
\end{equation}
\end{small}

{\bf CLIP} The GNN called Colored Local Iterative Procedure (CLIP) uses colors to disambiguate identical node features. Nodes with the same features are assigned to different subsets, and then the nodes in a same subset will be given a distinct color ${\bf c}_i$ from a finite color set $C$. And the colored graph can be represented as the following Equation. The CLIP is capable of capturing topology characteristics that traditional MPNNs fail to distinguish, and becomes a provably universal extension of MPNNs.
\begin{small}
\begin{equation}
G=({\bf A}, {\bf X} \oplus [{\bf c}_i]).
\end{equation}
\end{small}

{\bf RP-GNN} Murphy et al.\cite{murphy2019relational} considered randomly assigning an order of nodes as their extra features and proposed the relational pooling GNN (RP) as a general method in GNNs. This method takes advantage of the enumeration idea, by enumerating all possible results of different input orders, and then taking the average result as the final output. Suppose there are all $m$ possible permutations of graph $G$ and for one permutation $\pi$, ${\bf I}_{\pi}$ is a one-hot encode matrix of the permutation sequence, $\pi \circ G = ({\bf A}, {\bf X} \oplus {\bf I}_{\pi})$ denotes a permutated graph. Each permutated graph can get the node embeddings ${\bf h}^{\pi} _v$ via GNN, and take the average of them as the final embedding ${\bf h}_v$.
\begin{small}
\begin{equation}
{\bf h}_v = \frac{1}{m!} \sum_{m} \left( {\bf h}^{\pi} _v\right).
\end{equation}
\end{small}

{\bf RNI-GNN} The random node initialization (RNI)\cite{abboud2020surprising} is proposed as a means of training and implementing GNNs with randomized initial node features ${\bf R}$. RNI-GNNs are shown to be effective in achieving universal without relying on high-order feature calculations. This also holds with partially randomized initial node features.
\begin{small}
\begin{equation}
G=({\bf A}, {\bf X} \oplus {\bf R}).
\end{equation}
\end{small}

\subsubsection{Encoding the micro-topology}
Encoding the micro-topology is to determine the position of each node within the graph topology, accomplished through distance encoding or position encoding.

{\bf DE-GNN} The distance encoding (DE)\cite{li2020distance} is proposed as a generic initialization mean of contributing to the node embeddings in GNNs by capturing the distance from a given node to the node set to be learnt. Given a node set $S$ whose topology representation is to be learnt, for every node $u$ in the graph, DE is defined as a mapping $f_{\rm DE}$ of a set of landing probabilities of random walks from each node of the set $S$ to node $u$. DE-GNN is proved to be able to effectively distinguish node sets within nearly all regular graphs where traditional GNNs struggle. While it introduces significant memory and time overhead.
\begin{small}
\begin{equation}
{\bf d}_{v|S} = {\rm AGG}({f_{\rm DE}(v,u)|u \in S}),
\end{equation}
\end{small}
\begin{small}
\begin{equation}
{\bf h}_v ^{(0)} = {\bf x}_v \otimes {\bf d}_{v|S},
\end{equation}
\end{small}
where ${\bf d}_{v|S}$ denotes the distance of node $v$ and set $S$, $\otimes$ is the concatenation operation.

{\bf P-GNN} In order to mitigate the computational cost of DE, an approach called position encoding (PE) is applied using the absolute position of the nodes in the graph as an additional feature. This technique approximates DE by measuring the distances between position features and can be shared among different node sets. The position-aware GNN (P-GNN) proposed by You et al.\cite{you2019position} incorporates PE by measuring the distance from a node to a set of randomly selected anchor nodes. However, it has been observed that this approach has slow convergence rate and unsatisfactory performance. The procedure of PE is shown as following:
\begin{small}
\begin{equation}
{\bf d}_{v,u} ^{q}=\left\{\begin{array}{l}
{\bf d}_{v,u}, {\bf d}_{v,u} \leq q \\
\infty, \rm otherwise, 
\end{array}\right.
\label{equa14}
\end{equation}
\end{small}
\begin{small}
\begin{equation}
{\bf d}_{\rm standard}=\frac{1}{1+{\bf d}_{v,u} ^{q}},
\label{equa15}
\end{equation}
\end{small}
where $q$ is the hop from $v$ to $u$, ${\bf d}_{v,u}$ is the shortest distance of $v$ and $u$.

{\bf PEG} Utilizing PE, \cite{wang2022equivariant} introduces a novel approach to constructing more powerful GNNs - a class of GNN layers termed PEG. PEG uses separate channels to update both the node features and position features, exhibiting permutation equivariance and rotation equivariance simultaneously, where Stability is actually an even stronger condition than permutation equivariance because it characterizes how much gap between the predictions of the two graphs is expected when they do not perfectly match each other. 

In addition to DE or PE, incorporating marked topology information into message passing mechanisms shows significant enhancement of the GNNs expressive power.

{\bf SMP} The topology message passing (SMP)\cite{vignac2020building} method employs one-hot encoding to propagate the sole node identification, enabling the learning of local context matrices surrounding each node. This confers SMP with the capability to learn rich local topological information. In SMP, each node of a graph maintains a local context matrix ${\bf U}_i$ rather than a feature vector ${\bf x}_i$ as in MPNN. The ${\bf U}_i$ is initialized as a one-hot encoding ${\bf U}_i ^{(0)} = {\bf 1}$ for every $v_i \in \mathcal{V}$, and if there are features ${\bf x}_i$ associated with node $v_i$, they are appended to the same row of the local context as ${\bf U}_i ^{(0)} [i,:] = [1, {\bf x}_i]$. The message passing process is as Equation \ref{equa_SMP} showed.
\begin{small}
\begin{equation}
{\bf U}_i ^{(l+1)} = {\rm UPD}({\bf U}_i ^l, {\rm AGG}({{\rm MSG}({\bf U}_j ^l)}_{v_j \in N(v_i)}).
\label{equa_SMP}
\end{equation}
\end{small}

{\bf GD-WL} One of the main drawbacks of 1-WL is the lack of distance information between nodes. Inspired by some distance coding methods~\cite{you2019position, wang2022equivariant, vignac2020building}, \cite{zhang2023rethinking} further introduced a new color refinement framework called Generalized Distance Weisfeller-Lehman (GD-WL).The update rule of GD-WL is very simple and can be written as:
\begin{small}
\begin{equation}
{\bf c}_{v}^{(l)}={\rm Hash}\left({\bf c}_{v}^{(l-1)}, \left\{\left\{\left(d_{vu},{\bf c}_{u}^{(l-1)}\right(| u \in N(v)\right\}\right\}\right),
\end{equation}
\end{small}
where $d_{vu}$ can be an arbitrary distance metric.

\subsubsection{Encoding the local-topology}
This method focuses on local topology, mainly utilizing subgraphs explicitly or implicitly for node embedding to learn more topology information.

{\bf GSN} Graph sub-topology Networks (GSN)\cite{bouritsas2022improving} represent a topologically-aware message passing scheme based on sub-topology encoding. GSN explicit encodes the count of pre-defined subgraph topology appearance in the graph, as a topology role in message passing to capture a richer topological features and thereby exhibit superior expressive power over the WL test. However, the pre-defined subgraphs require the domain expert knowledge, hampering the model's widespread adoption. Considering ${\bf r}_v$ as the topology role induced by the subgraph counts, the message passing process is as following:
\begin{small}
\begin{equation}
{\bf m}^{(l+1)} _v = {\rm MSG}({{\bf h}^{(l)} _v,{\bf h}^{(l)} _u,{\bf r}_v,{\bf r}_u:u \in N(v)}).
\label{equa_gsn}
\end{equation}
\end{small}

{\bf GraphSNN} Wijesinghe et al.\cite{wijesinghe2021new} introduced the neighborhood subgraph graph neural network (GraphSNN), which enables the transmission of diverse messages based on the topology features of the central node's 1-hop neighborhood subgraph during message passing. The weight of the delivered message is determined by the extent of overlap subgraph between the central node and its neighbors in Equation \ref{equa5} and Equation \ref{equa6}, with edge overlap also factored into the weight calculation. By injecting the topology information of the neighborhood subgraph in the message passing process as Equation \ref{equa7}, GraphSNN can improve its expressive power without necessitating specialized knowledge or sacrificing computational complexity.
\begin{small}
\begin{equation}
w_{v u}=\frac{\left|E_{v u}\right|}{\left|V_{v u}\right| \cdot\left|V_{vu}-1\right|} \cdot\left|V_{v u}\right|^{\lambda},
\label{equa5}
\end{equation}
\end{small}
\begin{small}
\begin{equation}
{\bf W}=(\frac{w_{v u}}{\sum_{u \in N(v)} w_{v u}})_{v,u \in V},
\label{equa6}
\end{equation}
\end{small}
\begin{small}
\begin{equation}
\begin{split}
    h_{v}^{(l+1)} = {\rm MLP}_{\theta} \left(\gamma^{(l)}\left(\sum_{u \in N(v)} {\bf W}_{v u}+1\right){\bf h}_{v}^{(l)} \right.\\
    \phantom{=\;\;}
    \left.+\sum_{u \in N(v)}\left({\bf W}_{v u}+1\right){\bf h}_{u}^{(l)}\right),
\end{split}
\label{equa7}
\end{equation}
\end{small}
where $\lambda>0$ and the overlap subgraph is denoted as $\mathcal{G}_{v u}=\left(V_{v u}, E_{v u}\right)=S_{v} \cap S_{u}$ for neighboring nodes $v$, $u$. ${\bf W}$ is the weight matrix. $\gamma$ and $\theta$ is learnable scalar parameter. 

{\bf GNN-AK} \cite{zhao2021stars} extends the scope of local aggregation in GNNs beyond 1-hop neighbors (star-like) into more generalized subgraphs. This method, termed as GNN-AK, computes the encoding of induced subgraphs of nodes and serves as a wrapper to uplift any GNN, where GNN perform as a kernel. Let $N_k (v)$ be the set of nodes in the k-hop egonet rooted at node $v$, containing node $v$. For $N_k (v) \subseteq \mathcal{V}$, $G[N_k (v)]$ is the induced subgraph of $v$: $G[N_k (v)] = (N_k (v), {(i,j) \in \mathcal{E} | i,j \in N_k (v)}$. Every subgraph $G[N_k (v)]$ is encoded as ${\bf g}_{sub(v)}$, concatenating with the node embedding ${\bf h}_{sub(v)}$ aggregated from $N_k (v)$ , to obtain the final embedding ${\bf h}_v$:
\begin{small}
\begin{equation}
{\bf g}_{{\rm sub}(v)} ^{(l+1)}= {\rm POOL}^{(l)} ({{\bf h}_{{\rm sub}(i)} | i \in N_k (v)}),
\end{equation}
\end{small}
\begin{small}
\begin{equation}
{\bf h}_v = {\rm CON}({\bf g}_{{\rm sub}(v)}, {\bf h}_{{\rm sub}(v)}).
\end{equation}
\end{small}

{\bf NGNN} Similar to the design philosophy of GNN-AK, Nested Graph Neural Networks (NGNNs)\cite{zhang2021nested} extract a local subgraph around each node and apply a GNN to learning the representation of each subgraph. The overall graph representation is obtained by aggregating these subgraph representations. However, NGNN differs from GNN-AK in subgraph design. Specifically, the height of the rooted subgraph in NGNN can be determined based on the given hop of the node.

{\bf MPSN} The Message Passing Simplex Networks (MPSNs)\cite{bodnar2021weisfeiler}, which operates message passing on simplicial complexes (SCs)\cite{nanda2021computational}. SCs are in correspondence with subgraph topology, where SCs of varying dimensions concretize topology information of graph at different levels. Specifically, 0-simplices can be interpreted as vertices, 1-simplices as edges, 2-simplices as triangles, and so on.

{\bf ESAN} An empirical observation is that although two graphs may not be discriminable by MPNNs, they generally contain distinguishable subgraphs. Thus, \cite{bevilacqua2021equivariant} recommends representing each graph as a set of subgraphs derived from a predefined policy and proposes a novel equivariant subgraph aggregation network (ESAN) framework for handling them, which comprises the DSS-GNN architecture and subgraph selection policy. The core of the approach is to represent the graph $G$ as a bag (multiset) $S_G=\{\{G_1, ..., G_m\}\}$ of its subgraphs, and to make a prediction on a graph based on this subset. Motivated by the linear characterization, DSS-GNN adopt the same layer topology:
\begin{small}
\begin{equation}
F_{DSS-GNN} = E_{\rm sets} \circ R_{\rm subgraphs} \circ E_{\rm subgraphs},
\end{equation}
\end{small}
\begin{small}
\begin{equation}
E_{\rm subgraphs}: (L({\bf A},{\bf X}))_i = L^1 ({\bf A}_i,{\bf X}_i) + L^2 (\sum_{j=1} ^m {\bf A}_j,\sum_{j=1} ^m {\bf X}_i),
\end{equation}
\end{small}
where $E_{\rm subgraphs}$ is an equivariant feature encoding layer, $R_{\rm subgraphs}$ is a subgraph readout layer, $E_{\rm sets}$ is a universal set encoding layer. And $(L({\bf A},{\bf X}))_i$ denotes the output of the layer on the i-th subgraph, $L^1,L^2$ represent two graph encoders and can be any type of GNNs layer.

The methods outlined above pertain primarily to explicit utilization of subgraph topology, however implicit utilization can also serve to enhance the learning of graph topology.

{\bf LPR-GNN} Inspired by RP-GNN\cite{murphy2019relational} and subgraph counting task, the Local Relational Pooling model (LRP-GNN)\cite{chen2020can} is proposed to achieve competitive performance on sub-topology counting. LRP-GNN refers to RP-GNN that operates on subgraphs essentially, but in comparison to previous works~\cite{bouritsas2022improving, wijesinghe2021new, zhao2021stars}, LRP can not only compute sub-topology, but also learn they relevant relation of sub-topology. 

{\bf k-GNN} The higher-order message passing (k-GNN) scheme, implicitly utilizes subgraphs. The k-WL test serves as the foundation for k-GNNs. Whereas the standard WL test iteratively aggregates neighbors of a single node, k-WL operates on tuples of k-nodes for iterative aggregation. These k-tuples can be interpreted as subgraphs, where higher-order message passing occurs between subgraphs. Equation \ref{equa_N(s)} redefines the neighborhood representation of k-tuples, and Equation \ref{equa_kgnn} enables reliable capture of inherent local graph topology through information propagation and embedding update over k-neighborhoods. It has been proved that the discrimination power of 1-WL is equivalent to 2-WL. Moreover, for k$\geq$2, (k+1)-WL is strictly more powerful than k-WL. Notably, 3-WL is considered to be the strongest expressive power to identify all non-isomorphic graphs to date~\cite{arvind2020weisfeiler, morris2020weisfeiler}. The k-GNNs expressive power is equal to k-WL, hence when k$\geq$2, k-GNNs offers the strongest expression rivaled to 3-WL.
\begin{small}
\begin{equation}
N(s)=\left\{t \in[V]^{k} \| s \cap t \mid=k-1\right\},
\label{equa_N(s)}
\end{equation}
\end{small}
\begin{small}
\begin{equation}
{\bf h}_{k}^{(l+1)}(s)=\sigma\left({\bf h}_{k}^{(l)}(s) \cdot {\bf W}_{1}^{(l+1)}+\sum_{u \in N(s)} {\bf h}_{k}^{(l)}(u) \cdot {\bf W}_{2}^{(l+1)}\right),
\label{equa_kgnn}
\end{equation}
\end{small}
where all subsets of size $k$ over the set $\mathcal{V}$ are considered as $[V]^k$, and a $k$-set in $[V]^k$ is denoted by $s=\left\{v_{1}, \ldots, v_{k}\right\}$ as the central tuple. Moreover, $N(s)$ denotes the neighborhood of $s$, and ${\bf W}_{1}$, ${\bf W}_{2}$ are the parameter matrix.

{\bf K-hop GNN} In MPNN, the representation of a node is updated by aggregating its direct neighbours, which are called 1-hop neighbours. Some work extends the message passing to the concept of ~\cite{abu2019mixhop, nikolentzos2020k, wang2020multi, chien2020adaptive, brossard2020graph, ying2021transformers} K-hops. In K-hop message passing, the node representation is updated not only by aggregating the information of the node's 1-hop neighbours, but also by aggregating all the neighbours within the node's K-hop. The K-hop neighbours of node $v$ refer to all neighbours whose distance from node $v$ is less than or equal to K. Denote by ${N(v)}^k$ the K-th hop neighbour of node $v$, i.e., the neighbour whose distance to the node is exactly equal to K.
\begin{small}
\begin{equation}
{\bf m}_{v}^{(l,k)}= {\rm MSG}^{(l)}\left(\left\{\left\{{\bf h}_{u}^{(l-1)}: u \in {N(v)}^k \right\}\right\}\right),
\end{equation}
\end{small}
\begin{small}
\begin{equation}
{\bf h}_{v}^{(l,k)}={\rm UPD}^{(l)}\left({\bf m}_{v}^{(l,k)},{\bf h}_{v}^{(l-1)}\right),
\end{equation}
\end{small}
\begin{small}
\begin{equation}
{\bf h}_{v}^{(l)}= {\rm AGG}^{(l)}\left(\left\{\left\{{\bf h}_{v}^{(l,k)}| k=1,2,...,K \right\}\right\}\right).
\end{equation}
\end{small}

The K-hop message passing is more powerful than 1-WL and can distinguish almost all rule-graphs, but its expressive power is similarly limited by 3-WL.

{\bf KP-GNN} introduced K-hop peripheral-subgraph-enhanced GNN (KP-GNN)\cite{feng2022powerful}, a new GNN framework with k-hop message passing, which significantly improves the expressive power of K-hop GNN. KP-GNN aggregates in the neighbour aggregation of each hop not only the neighbouring nodes, but also aggregates peripheral subgraphs (subgraphs induced by neighbours in that hop). This additional information helps KP-GNN learn more expressive local topology information around nodes. This improvement occurs in the message passing process:
\begin{small}
\begin{equation}
{\bf m}_{v}^{(l,k)}= {\rm MSG}^{(l)}\left(\left\{\left\{{\bf h}_{u}^{(l-1)}: u \in {N(v)}^k \right\}\right\}, G_v^k\right),
\end{equation}
\end{small}
where $G_v^k$ is subgraphs induced by neighbours of $v$ in hop $k$.

{\bf N2-FWL} extends the traditional k-FWL algorithm by constructing a neighbourhood tuple\cite{feng2024extending}. This neighbourhood tuple contains all possible combinations of node substitutions, which enables to capture richer information about the graph structure. By considering a wider range of neighbourhood information, N2-FWL is able to capture higher-order structures in the graph, such as 6-rings, 4-connected subgraphs, and 4-paths, which improves the expressive power of the model. In N2-FWL, the color update rule for node pair $(v_1, v_2)$ can be expressed as:
\begin{small}
\begin{equation}
\begin{split}
    \hat{C}_l^{(2,2)}\left(v_1,v_2\right)&={\rm HASH}\left({\hat{C}_{l-1}^{(2,2)}\left(v_1, v_2\right),}\right.\\
    &\left. {\left\{\hat{C}_{l-1}^{(2,2)}\left(u_1,u_2\right) \mid\left(u_1, u_2\right) \in \mathcal{Q}_F^{\left(w_1, w_2\right)}\left(v_1, v_2\right)\right\}}\right),
\end{split}
\end{equation}
\end{small}
where $\mathcal{Q}_F^{\left(w_1, w_2\right)}\left(v_1, v_2\right)$ denotes the neighbourhood tuple based on the node pairs $(\mathrm{w} 1, \mathrm{w} 2)$, and $\hat{C}_l^{(2,2)}$ denotes the color of the node pairs in the first iteration. N2-FWL is able to capture the higher-order structure of graphs, has theoretically comparable expressive power to 3-FWL, and can be designed with different instances according to different task requirements. However, it may face the problem of computational complexity in practical applications, especially when dealing with large graphs.

{\bf PIN} The Path Isomorphism Network (PIN)\cite{truong2024weisfeiler} implements the Path Weisfeiler-Lehman (PWL) test by lifting the graph-to-path complex, using simple paths as base elements, and updating the feature representation based on part-whole and adjacency relationships through a message passing process. This approach does not rely on the assumption of graph substructures (e.g., clusters, rings, etc.), but focuses on simple paths of the graph. In this way PIN enhances the understanding of the graph structure and thus improves its expressive power. In the PWL test, for each time step $t$ and each simple path $\sigma$, a hash function is used to update the colors based on the current color and the associated set of colors:
\begin{small}
\begin{equation}
ct_\sigma^{(t+1)}={\rm HASH}\left(ct_\sigma^{(t)}, ct_{B(\sigma)}^{(t)}, ct_{\uparrow(\sigma)}^{(t)}, ct_{\downarrow(\sigma)}^{(t)}\right),
\end{equation}
\end{small}
where $c t_\sigma^{(t)}$ is the color of the path $\sigma$ at time step $t$, $B(\sigma)$ is the set of boundaries of $\sigma$, and $\uparrow(\sigma)$ and $\downarrow(\sigma)$ are the set of up-neighbours and down-neighbours of $\sigma$, respectively. The final color of the path complex $P$ is represented using the set of colors of all simple paths.

\subsubsection{Encoding the global-topology}
Encoding the global-topology is to directly extract the topology information of the whole graph for learning.

{\bf Eigen-GNN} Zhang et al.\cite{zhang2021eigen} introduced the topology eigen decomposition module, designed as a universal plug-in to GNNs termed as Eigen-GNN. This approach incorporates eigen decomposition of the adjacency matrix ${\bf A}$ of the graph to obtain eigenvectors as topology features, which are subsequently concatenated with node features. The resulting amalgamation is then utilized as input to GNNs, as denoted in Equation \ref{equa_eigen}.
\begin{small}
\begin{equation}
{\bf H}^{(0)}=[{\bf X}, f({\bf Q})],
\label{equa_eigen}
\end{equation}
\end{small}
where ${\bf Q}$ is the eigenvectors corresponding to the top-d largest absolute eigenvalues of ${\bf A}$, $f$ is a function such as normalization or identity mapping.

{\bf SAGNN} Zeng et al.\cite{zeng2023substructure} enhances the model's ability to perceive the graph structure by innovatively introducing the subgraph encoding and information injection mechanism, which captures the local structural information around the nodes by Ego subgraph and the global structural information of the whole graph by Cut subgraph, and then injects the structural information encoded by the subgraph into the message passing process of the GNNs. SAGNN obtains theoretically more powerful expressive power than 1-WL, but may face the problem of higher computational cost in practical applications, especially when dealing with large graphs.

\subsection{GNNs architecture enhancement}
\subsubsection{Improving aggregation functions}
Improving aggregation functions involves improvements to the aggregation operations, wherein most of the modifications are made while retaining the permutation invariance of the aggregation function. However, there exist a few exceptions to break the permutation invariance. The improvement of aggregator that preserves invariance is established based on the theory proposed by Xu et al.\cite{xu2018powerful}.

{\bf GIN} The essence of MPNN is actually a function that maps multiple set, where allow repeated elements to exist. The capacity of multi-set functions to discern these repeated elements inherently constrains the expressive power of MPNN. As a result, optimizing the aggregation function necessitates the development of an injective function operating on multisets, ensuring distinct embeddings for different nodes. The graph isomorphism network (GIN) is a simple architecture to satisfy this prerequisite:
\begin{small}
\begin{equation}
{\bf h}_{v}^{(l+1)}={\rm MLP}\left(\left(1+\varepsilon^{(l)}\right) {\bf h}_{v}^{(l)}+\sum_{u \in N(v)} {\bf h}_{u}^{(l)}\right),
\end{equation}
\end{small}
where $\varepsilon^{(l)}$ represents the learnable or fixed scalar and the MLP plays a critical role in strengthening the expressive power by ensuring the injective property.

{\bf modular-GCN} The global parameter assignment strategy enforced by GCN is actually a permutation invariant aggregation operation. Dehmamy et al.\cite{dehmamy2019understanding} has insight into the loss of topology information in GCN, and designed three GCN modules using three different propagation strategies as (1)$f_{1}={\bf A}$, (2)$f_2={\bf D}^{-1}{\bf A}$, and (3)$f_3={\bf D}^{-1/2} {\bf A} {\bf D}^{-1/2}$ to address this limitation. Combining the three GCN modules compensates for the loss of missing topology information, realizing the optimization of the aggregation function.

Furthermore, the general aggregation operations in GNNs are interpreted as graph convolutions. The graph topology acts as a node feature filter, smoothing node features during the convolution process, which corresponds to a filtering operation. Thereby enable the analysis of GNNs expressive power through spectral decomposition techniques\cite{balcilar2021analyzing}.

{\bf diagonal-GNN} Kanatsoulis et al.\cite{kanatsoulis2022graph} utilizes spectral decomposition tools to view GNNs as graph feature filters, as shown in Equation \ref{equa_filter}, and proposed a GNN designed with the diagonal modules (diagonal-GNN). The use of diagonal modules allows GNNs to capture broader topology relationships while also capturing neighborhood relationships, enabling GNNs show the equivalent recognition abilities of featured graphs even on unfeatured graphs. 
\begin{small}
\begin{equation}
{\bf H} = \sigma(\sum_{k=0} ^{K-1} {{\bf A}^k {\bf X} {\bf W}_k}),
\label{equa_filter}
\end{equation}
\end{small}
\begin{small}
\begin{equation}
diagonal-GNN: {\bf H} = \sigma(\sum_{k=0} ^{K-1} {diag({\bf A}^k) {\bf X} {\bf W}_k}),
\end{equation}
\end{small}
where $K$ is the length of graph filter, $\sigma$ denotes a nonlinearity, ${\bf W}_k$ is the filter parameters and can be a matrix, a vector, or a scalar.

{\bf GNN-LF/HF} Leveraging spectral decomposition tools to design various convolution processes to handle specific filtering conditions, GNNs can recognize and process more complex low-frequency or high-frequency information. \cite{zhu2021interpreting} considers two distinct modes of low-pass filtering (LF) and high-pass filtering (HF), develop corresponding convolution operations, and design new models GNN-LF/HF respectively. The propagation mechanism of GNN-LF can be expressed by the following equations:
\begin{small}
\begin{equation}
	\begin{split}
		{\bf H}^{(l+1)}&=f_{\rm COM}(\frac{1+\alpha \mu-2 \alpha}{1+\alpha \mu-\alpha} \tilde{{\bf D}}^{-1 / 2} \tilde{{\bf A}} \tilde{{\bf D}}^{-1 / 2} {\bf H}^{(l)}\\
		&+\frac{\alpha \mu}{1+\alpha \mu-\alpha} {\bf X}+\frac{\alpha-\alpha \mu}{1+\alpha \mu-\alpha} \tilde{{\bf D}}^{-1 / 2} \tilde{{\bf A}} \tilde{{\bf D}}^{-1 / 2} {\bf X}),
	\end{split}
\end{equation}
\end{small}
\begin{small}
\begin{equation}
	{\bf H}^{(0)}=\frac{\mu}{1+\alpha \mu-\alpha} {\bf X}+\frac{1-\mu}{1+\alpha \mu-\alpha} \tilde{{\bf D}}^{-1 / 2} \tilde{{\bf A}} \tilde{{\bf D}}^{-1 / 2} {\bf X}.
\end{equation}
\end{small}

The propagation mechanism of GNN-HF is given by the following equations:
\begin{small}
\begin{equation}
	\begin{split}
		{\bf H}^{(l+1)}&=f_{\rm COM}(\frac{\alpha \beta-\alpha+1}{\alpha \beta+1} \tilde{{\bf D}}^{-1 / 2} \tilde{{\bf A}} \tilde{{\bf D}}^{-1 / 2} {\bf H}^{(l)}\\
		&+\frac{\alpha}{\alpha \beta+1} {\bf X}+\frac{\alpha \beta}{\alpha \beta+1} \tilde{{\bf D}}^{-1 / 2} \tilde{{\bf A}} \tilde{{\bf D}}^{-1 / 2} {\bf X}),
	\end{split}
\end{equation}
\end{small}
\begin{small}
\begin{equation}
	{\bf H}^{(0)}=\frac{1}{\alpha \beta+1} {\bf X}+\frac{\beta}{\alpha \beta+1} \tilde{{\bf D}}^{-1 / 2} \tilde{{\bf A}} \tilde{{\bf D}}^{-1 / 2} {\bf X},
\end{equation}
\end{small}
where $\mu$, $\alpha$, $\beta$ are hyper parameters that control the trade-off between the topology and feature information at different scales. $\tilde{{\bf A}}$ and $\tilde{{\bf D}}$ are the normalized adjacency matrix and the diagonal degree matrix, respectively, that account for the topology of the graph data.

{\bf Geom-GCN} Pei et al.\cite{pei2020geom} leverage the network geometry theory to propose a geometric aggregation scheme, implemented within graph convolution networks as Geom-GCN. This scheme maps graphs to a continuous latent space via node embeddings and subsequently constructs topology neighborhoods based on the geometric relations defined in the latent space, effectively preserving the intricate topological patterns present within the graphs. Building upon the definition of the original neighborhood, they further define topology neighborhoods $\mathcal{N}(v)=(\{N_g (v),N_s (v)\},\tau)$, consisting of a set of neighborhood $\{N_g (v),N_s (v)\}$ and a relation operator on neighborhoods $\tau$. $N_g (v)=\{u|u \in \mathcal{V},(u,v)\in \mathcal{E}\}$ is the neighborhood in the graph. $N_s (v)=\{u|u \in \mathcal{V},d({\bf z}_u,{\bf z}_v)<\rho\}$ is the neighborhood in the latent space, where $d$ is the distance function, $\rho$ is the pre-given parameter, and ${\bf z}_v \in {\bf R}$ can be considered as the position of nodes in the latent space via the mapping $f:v \rightarrow {\bf z}_v$. And $\tau:({\bf z}_v, {\bf z}_u) \rightarrow r \in R$, where $R$ is the set of the geometric relationships and $r$ can be specified as an arbitrary geometric relationship of interest according to the particular latent space and applications. On the basis, a bi-level aggregation scheme is introduced:
\begin{small}
\begin{equation}
{\bf m}_{(i,r)} ^{v,l+1} = \mathop{\rm AGG_1}\limits _{i \in {g,s}} (\{{\bf h}_u^l | u \in N_i (v), \tau({\bf z}_u, {\bf z}_u)=r\}),
\end{equation}
\end{small}
\begin{small}
\begin{equation}
{\bf m}_{v} ^{l+1} = \mathop{\rm AGG_2}\limits _{i \in {g,s}}(({\bf m}_{(i,r)} ^{v,l+1},(i,r))),
\end{equation}
\end{small}
where ${\bf m}_{(i,r)} ^{v,l+1}$ can be considered as the feature of virtual nodes in the latent space which is indexed by $(i, r)$ which is corresponding to the combination of a neighborhood $i$ and a relationship $r$.

The approach of optimizing the permutation invariant aggregation function presents significant limitations. Therefore, alternative designs have been proposed to address this constraint. 

{\bf PG-GNN} Permutation sensitive functions are regarded as a symmetry breaking mechanism that disrupts the equidistant status of adjacent nodes, allowing for the capture of relationships between neighboring nodes. However, few permutation-sensitive GNNs~\cite{hamilton2017inductive, murphy2018janossy, dasoulas2019coloring} are available to real-world applications, due to their complexity. Huang et.al.\cite{huang2022going} adopts Recurrent neural networks (RNNs) to model the dependency relationships between neighboring nodes as a permutation-sensitive function approach, wherein all permutations $\pi$ act on the $\mathcal{V}$, leading to the following aggregation strategy: 
\begin{small}
\begin{equation}
{\bf h}_v ^{(l+1)} = \sum_{\pi in \Pi} {\rm RNN}({\bf h}_{\pi(u_1)} ^{(l)},{\bf h}_{\pi(u_2)} ^{(l)},...,{\bf h}_{\pi(u_n)} ^{(l)}) + {\bf W}^{(l)}{\bf h}_v ^{(l)},
\end{equation}
\end{small}where $u_{1:n} \in N(v)$, ${\bf W}$ is the parameter matrix.

\subsubsection{Adopting equivariant architecture}
The bottleneck of GNNs expressive power lies in the permutation invariance of their aggregation operators\cite{loukas2019graph}. While the limitation only serves as a condition to induce permutation equivariance, rather than a requisite component in GNN design. Consequently, abandoning aggregation operations and employing distinct arrangements of equivariant functions or operators can lead to the direct design of equivariant GNNs~\cite{maron2018invariant, maron2019provably, geerts2020expressive, keriven2019universal, azizian2020expressive, chen2019equivalence, frasca2022understanding, balcilar2021breaking, geerts2022expressiveness}. However, it is worth noting that equivariant GNNs perform calculations on global features, demanding substantial memory and computational resources, rendering these networks primarily as theoretical tools with limited practical applicability. Hence, the following section will offer a succinct introduction to several equivariant GNNs models as a supplement to the method of improving GNNs expressive power.

{\bf k-IGN} Maron et.al.~\cite{maron2018invariant, maron2019universality} develop the k-order linear invariant graph neural network (k-IGN), which operates calculation directly on global information tensors. K-IGN comprises of a sequence of alternating layers composed of invariant or equivariant linear layers and nonlinear activation layers and demonstrates a strong expressive power not less than MPNNs, equivalent to the k-WL test\cite{geerts2020expressive}.

The provably powerful graph networks ({\bf PPGN})\cite{maron2019provably}, the folklore graph neural network ({\bf FGNN})\cite{geerts2020expressive}, \cite{azizian2020expressive} and the {\bf Ring-GNN}\cite{chen2019equivalence} are improvements to k-IGN, and by simplifying the computational process, they can achieve 3-WL expressive power at k=2.

{\bf SUN} Inspired by a series of subgraph GNNs~\cite{papp2021dropgnn, papp2022theoretical, abboud2020surprising, cotta2021reconstruction}, Frasca et al.\cite{frasca2022understanding} proposes a novel subgraph learning framework based on 2-IGN, termed as the Subgraph Union Network (SUN). Rather than improving the computational process, SUN improves its expression by optimizing the 2-IGN layer to capture local topology information.

The mentioned models are designed based on the IGN framework, which utilizes arrangements of equivariant functions to obtain equivariant networks. The following will introduce some designs that obtain equivariant networks by arranging equivariant operators~\cite{barcelo2020expressive, barcelo2020logical, grohe2021logic, geerts2022expressive, geerts2022expressiveness, balcilar2021breaking}.

These models are formulated using matrix language (ML) or tensor language (TL) to describe operators, which are then combined in various ways to obtain GNNs with different expressive powers. Balcilar et al.\cite{balcilar2021breaking} employ equivariant operators in ML to design two models, {\bf GNNML1} and {\bf GNNML3}, with 1-WL and 3-WL expressive powers, respectively. Meanwhile, Geerts et al.\cite{geerts2022expressiveness} utilize the ML to formulate {\bf k-MPNN}.

These higher-order equivariant networks are designed to improve the expressive power by the inspiration of the k-WL hierarchy. Ideally, we could improve the expression of the models by increasing k in the k-WL hierarchy. However, in fact, due to the combinatorial growth of complexity, these models can only be applied in practice~\cite{maron2019provably, maron2019universality, morris2019weisfeiler, morris2017glocalized} to small graphs with small k. Specifically the limit of practical application is reached at k = 3. Their poor scalability becomes a key limitation for the practical application of such networks.

\subsection{Comparison and analysis of Methods}
\subsubsection{Advantages and drawbacks of methods}
{\bf Graph feature enhancement} enhances the model's understanding of node/edge features, beneficial for tasks with complex feature relationships, like various tasks on heterogeneous graphs.

Extracting feature dependencies is suitable for tasks where feature relationships significantly influence performance. However, introducing complex dependencies can lead to overfitting and raise computational demands.

Enhancing feature utilization can improve the interpretability and accuracy of classification results, crucial for applications requiring a thorough understanding of feature distributions. However, the introduction of random features can disrupt permutation invariance, necessitating additional design considerations\cite{murphy2019relational} to maintain this property.

\noindent {\bf Graph topology enhancement} strengthens the model's comprehension of graph structures, particularly in sparse-feature contexts. However, it requires domain knowledge for feature selection, and potentially escalate memory and computational expenses, hindering scalability on large datasets.

Adding extra topology information is beneficial in sparse-feature scenarios, commonly used in social network analysis and recommendation systems, but often require domain knowledge for effective feature selection.

Encoding micro-topology is critical for tasks requiring precise, relative or absolute positional information, yet introduces high implementation complexity and additional memory and computational costs.

Encoding local-topology is effective for tasks dependent on local structural features, such as community detection and network security, but it can be computationally demanding and require expert knowledge.

Encoding global-topology is essential for tasks needing comprehensive graph understanding, such as graph classification and network traffic analysis, but overemphasis on global information risks information overload.

\noindent {\bf GNNs architecture enhancement} enables the model to learn complex graph structures more effectively. However, the high memory and computational demands restrict its application to large datasets, and the addition of hyper-parameters complicates the model.

Improving aggregation functions is suitable for tasks requiring specific topological pattern recognition and spectral analysis, but increases computational load and necessitates meticulous hyper-parameter tuning.

Adopting equivariant architectures provides robust theoretical expressivity, comparable to the k-WL test, often serving as a theoretical tool for understanding GNNs expressive power. However, due to their reliance on global feature computation, these architectures demand substantial memory and computational resources, limiting their scalability to large datasets.

\subsubsection{Assessment of some existing models}
Table \ref{table_node} and Table \ref{table_graph} present a comparative analysis of model performance across node classification and graph classification tasks using three real-world datasets separately: Cora, Citeseer, PubMed for node classification, and ZINC, MUTAG, PROTEINS for graph classification. Metrics include  MAE for graph classification on ZINC dataset and accuracy for other tasks. The analysis encompasses a range of GNN baselines, including GCN\cite{kipf2016semi}, GAT\cite{velivckovic2017graph}, and GraphSAGE\cite{hamilton2017inductive}, with experimental results sourced from scholarly literature.

This table indicates that  graph feature enhancement is primarily used for node classification, while graph topology enhancement is better suited for graph classification. GNN architecture enhancement is versatile for both tasks. Model performance varies significantly across in different tasks, with excellence in one not guaranteeing proficiency in others, highlighting the need for method selection tailored to dataset characteristics and task requirements. For node classification, improving aggregation functions improves expressiveness and accuracy. For graph classification, encoding global-topology is preferred. Additionally, computational efficiency and model scalability are crucial considerations in method choices.

\begin{table}[!t]
\renewcommand{\arraystretch}{1.3}
\caption{Performance of Various GNN Models on Node Classification Datasets. This table compares the performance of various GNNs on three different node classification datasets (Cora, Citeseer, PubMed), using accuracy as metric, categorized by methods that enhance the model's expressive power. Best: underlined; second-best: dashed lines.}
\label{table_node}
\centering
\resizebox{\linewidth}{!}{ 
\begin{tabular}{|c||c||c||c|}
\hline Model & Cora & Citeseer & PubMed\\

\hline 
GCN\cite{kipf2016semi} & $0.811 \pm 0.003$ & $0.701 \pm 0.004$ & $0.786 \pm 0.006$\\
GAT\cite{velivckovic2017graph} & $0.814 \pm 0.010$ & $0.709 \pm 0.007$ & $0.782 \pm 0.008$\\
GraphSAGE\cite{hamilton2017inductive} & $0.804 \pm 0.004$ & $0.696 \pm 0.003$ & $0.775 \pm 0.008$\\

\hline 
AM-GCN\cite{wang2020gcn} & - & \dashuline{$0.745 \pm 0.010$} & -\\
CL-GNN\cite{hang2021collective} & $0.748 \pm 0.001$ & - & $0.765 \pm 0.002$\\
MM-GNN\cite{Bi2023MMGNN:MG} & $0.842 \pm 0.006$ & $0.730 \pm 0.006$ & $0.803 \pm 0.007$\\

\hline 
ID-GNN\cite{you2021identity} & \underline{$0.870 \pm 0.220$} & $0.742 \pm 0.010$ & -\\
PEG\cite{wang2022equivariant} & $0.822 \pm 0.001$ & $0.719 \pm 0.001$ & $0.799 \pm 0.001$\\
Eigen-GNN\cite{zhang2021eigen} & $0.789 \pm 0.007$ & $0.665 \pm 0.003$ & $0.786 \pm 0.001$\\
GraphSNN\cite{wijesinghe2021new} & $0.838 \pm 0.012$ & $0.735 \pm 0.016$ & $0.796 \pm 0.014$\\
\hline 

GIN\cite{xu2018powerful}  & $0.768 \pm 0.009$ & $0.677 \pm 0.003$ & $0.770 \pm 0.005$\\
GNN-LF\cite{zhu2021interpreting} & $0.837 \pm 0.001$ & $0.720 \pm 0.003$ & $0.803 \pm 0.002$\\
GNN-HF\cite{zhu2021interpreting} & $0.840 \pm 0.002$ & $0.723 \pm 0.003$ & \dashuline{$0.804 \pm 0.003$}\\
Geom-GCN\cite{pei2020geom} & \dashuline{$0.854 \pm 0.004$} & \underline{$0.760 \pm 0.015$} & \underline{$0.879 \pm 0.025$}\\
\hline 

\end{tabular}
}
\end{table}

\begin{table}[!t]
\renewcommand{\arraystretch}{1.3}
\caption{Performance of Various GNN Models on Graph Classification Datasets. This table compares the performance of various GNNs on three different graph classification datasets (ZINC, MUTAG, PROTEINS), using MAE(ZINC) and accuracy(MUTAG, PROTEINS) as metrics, categorized by methods that enhance the model's expressive power. Best: underlined; second-best: dashed lines.}
\label{table_graph}
\centering
\resizebox{\linewidth}{!}{ 
\begin{tabular}{|c||c||c||c|}
\hline Model & ZINC & MUTAG & PROTEINS\\

\hline 
GCN\cite{kipf2016semi} & $0.321 \pm 0.009$ & $0.891 \pm 0.058$ & $0.752 \pm 0.051$\\
GAT\cite{velivckovic2017graph} & $0.384 \pm 0.007$ & $0.901 \pm 0.058$ & $0.759 \pm 0.043$\\
GraphSAGE\cite{hamilton2017inductive} & $0.384 \pm 0.007$ & $0.851 \pm 0.076$ & $0.692 \pm 0.100$\\

\hline 
Twin-GNN\cite{wang2022twin} & - & $0.900 \pm 0.039$ & -\\
ID-GNN\cite{you2021identity} & $0.083 \pm 0.003$ & $0.904 \pm 0.054$ & $0.754 \pm 0.027$\\
CLIP\cite{dasoulas2019coloring} & - & $0.939 \pm 0.040$ & $0.771 \pm 0.044$\\
GD-WL\cite{zhang2023rethinking} & $0.081 \pm 0.009$ & - & -\\
SAGNN\cite{zeng2023substructure} & $0.072 \pm 0.002$ & \dashuline{$0.952 \pm 0.030$}  & \dashuline{$0.798 \pm 0.038$} \\
GSN\cite{bouritsas2022improving} & $0.101 \pm 0.010$ & $0.922 \pm 0.075$ & $0.766 \pm 0.050$\\
GraphSNN\cite{wijesinghe2021new} & - & $0.947 \pm 0.019$ & $0.784 \pm 0.027$\\
GNN-AK\cite{zhao2021stars} & $0.105 \pm 0.010$ & $0.913 \pm 0.070$ & $0.771 \pm 0.057$\\
NGNN\cite{zhang2021nested} & $0.111 \pm 0.003$ & $0.879 \pm 0.082$ & $0.742 \pm 0.037$\\
MPSN\cite{bodnar2021weisfeiler} & $0.079 \pm 0.006$ & - & $0.764 \pm 0.033$\\
ESAN\cite{bevilacqua2021equivariant} & $0.102 \pm 0.003$ & $0.911 \pm 0.070$ & $0.759 \pm 0.043$\\
LRP-GNN\cite{chen2020can} & $0.223 \pm 0.008$ & - & -\\
3-GNN\cite{morris2019weisfeiler} & $0.407 \pm 0.028$ & $0.861 \pm 0.001$ & $0.755 \pm 0.001$\\
3-hop GNN\cite{nikolentzos2020k} & - & $0.876 \pm 0.007$ & $0.753 \pm 0.004$\\
KP-GNN\cite{feng2022powerful} & - & \underline{$0.961 \pm 0.046$} & \underline{$0.801 \pm 0.043$}\\
N2-FWL\cite{feng2024extending} & \dashuline{$0.059 \pm 0.001$}  & - & -\\
PIN\cite{truong2024weisfeiler} & $0.096 \pm 0.003$ & - & $0.788 \pm 0.044$\\
\hline 

GIN\cite{xu2018powerful} & $0.387 \pm 0.015$ & $0.894 \pm 0.056$ & $0.762 \pm 0.028$\\
PG-GNN\cite{huang2022going} & $0.282 \pm 0.011$ & - & $0.768 \pm 0.038$\\
IGN\cite{maron2018invariant} & \underline{$0.045 \pm 0.006$} & $0.839 \pm 0.130$ & $0.766 \pm 0.055$\\
PPGN\cite{geerts2020expressive} & $0.159 \pm 0.007$ & $0.906 \pm 0.087$ & $0.772 \pm 0.047$\\
Ring-GNN\cite{chen2019equivalence} & $0.353 \pm 0.019$ & $0.868 \pm 0.064$ & $0.757 \pm 0.029$\\
SUN\cite{frasca2022understanding} & $0.083 \pm 0.003$ & $0.727 \pm 0.058$ & $0.758 \pm 0.050$\\
GNNML1\cite{balcilar2021breaking} & $0.314 \pm 0.015$ & $0.900 \pm 0.043$ & $0.758 \pm 0.049$\\
GNNML3\cite{balcilar2021breaking} & $0.161 \pm 0.006$ & $0.909 \pm 0.055$ & $0.764 \pm 0.051$\\
k-MPNN\cite{geerts2022expressiveness} & $0.145 \pm 0.007$ & - & -\\
\hline 

\end{tabular}
}
\end{table}

\section{Challenge and future directions}\label{sec5}
\textbf{Refinement of assessment indicators.} Current research, such as by Zhang et al.\cite{zhang2024beyond}, has introduced quantitative criteria to assess GNN expressive power, an improvement over the initial WL tests. These criteria enhance model scalability and reduce abrupt task shifts\cite{Hawkins2004ThePO}. However, they only cover topological representation, neglecting feature embedding ability, which is known to interact positively with topology. Thus, there's a need for further refinement of these metrics to include comprehensive evaluation of both feature embedding and topological representation, with the expectation of seeing such metrics developed soon.

\textbf{More GNNs Architecture Design.} Message passing in GNNs has seen widespread success\cite{VELICKOVIC2023102538} but is limited by the WL test in expressive power. Alternatives like LGNN face development hurdles due to tensor calculation complexities. Future research should focus on designing more expressive GNN architectures, possibly incorporating multi-level messaging and multi-scale graph embedding to capture complex graph features. Integrating GNNs with physical models, such as molecular dynamics for improved molecular structure and property capture, is also promising. Work by Bronstein et al.\cite{bouritsas2022improving} on a physics-inspired continuous learning GNNs model that uses differential geometry and algebraic topology is a step towards overcoming MPNN paradigm limitations.

\textbf{Graph Transformer Modeling.} Transformers, a special case of GNNs\cite{VELICKOVIC2023102538}, enhance graph data processing by incorporating topological information, offering superior capability in capturing complex node relationships and addressing missing elements and long-distance dependencies. This approach potentially reduces over-smoothing and over-squashing issues inherent in GNNs. The integration of Transformer architecture is anticipated to foster the development of Graph Transformers with enhanced expressive power, possibly outperforming traditional GNNs and expanding their application scope.

\section{Conclusion}\label{sec6}
GNNs research has seen substantial growth, yielding models with enhanced expressive power. Yet, a deeper grasp of this expressivity remains elusive. To advance understanding, we introduce a comprehensive theoretical framework that defines, explores the limitations, and analyzes factors influencing GNNs expressive power. This framework also organizes current methods for boosting expressiveness. As it encompasses both graph features and topology, our approach provides a systematic methodology for studying GNNs expressive power, highlighting the need for methods that consider these dual aspects.

\section*{Acknowledgments}
This study was supported by the National Natural Science Foundation of China (NSFC, Grant Number: 72421002, 72025405, 62206303, U23A20296, 62272469) and Science and Technology Innovation Program of Hunan Province (Grant Number: 2023RC3009, 2023RC1007)

\normalem
\bibliographystyle{plain}
\bibliography{ref}

\begin{IEEEbiography}[{\includegraphics[width=1in,height=1.25in,clip,keepaspectratio]{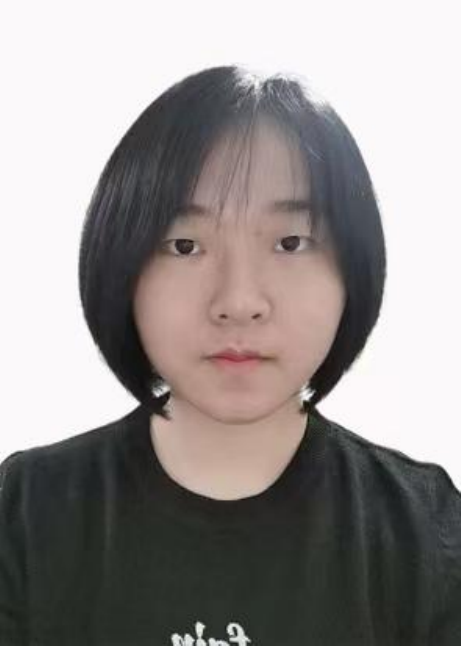}}]{Bingxu Zhang}
received the B.S. degree in Big Data Engineering from National University of Defense Technology, Changsha, China,in 2021. She is currently working toward the Ph.D. degree in Management Science and Engineering with the School of Systems Engineering,National University of Defense Technology, Changsha, China. Her research interests include graph neural networks and combinatorial optimization.
\end{IEEEbiography}

\vspace{-12mm}
\begin{IEEEbiography}[{\includegraphics[width=1in,height=1.25in,clip,keepaspectratio]{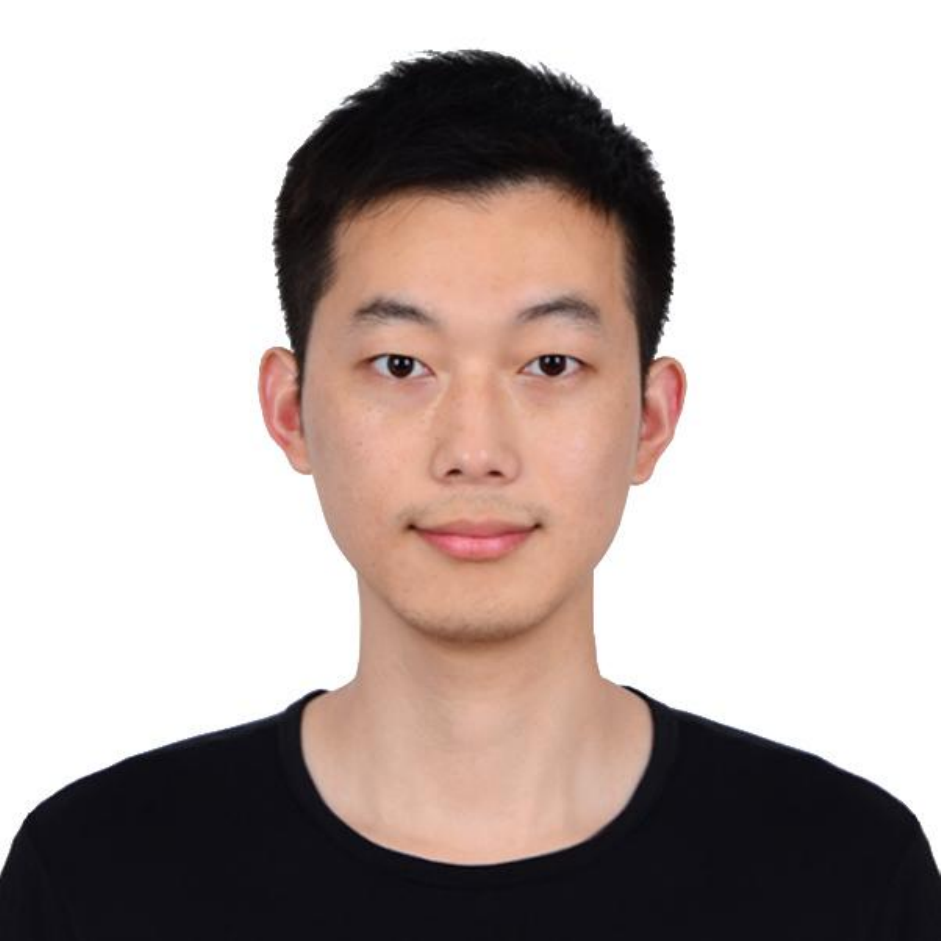}}]{Changjun Fan}
received the B.S. degree, M.S. degree and PhD degree all from National University of Defense Technology, Changsha, China, in 2013, 2015 and 2020. He is also a visiting scholar at Department of Computer Science, University of California, Los Angeles, for two years.  He is currently an associate professor at National University of Defense Technology. His research interests include deep graph learning and complex systems, with a special focus on their applications on intelligent decision making. During his previous study, he has published a number of refereed journals and conference proceedings, such as Nature Machine Intelligence, Nature Communications, AAAI, CIKM, etc. 
\end{IEEEbiography}

\vspace{-12mm}
\begin{IEEEbiography}[{\includegraphics[width=1in,height=1.25in,clip,keepaspectratio]{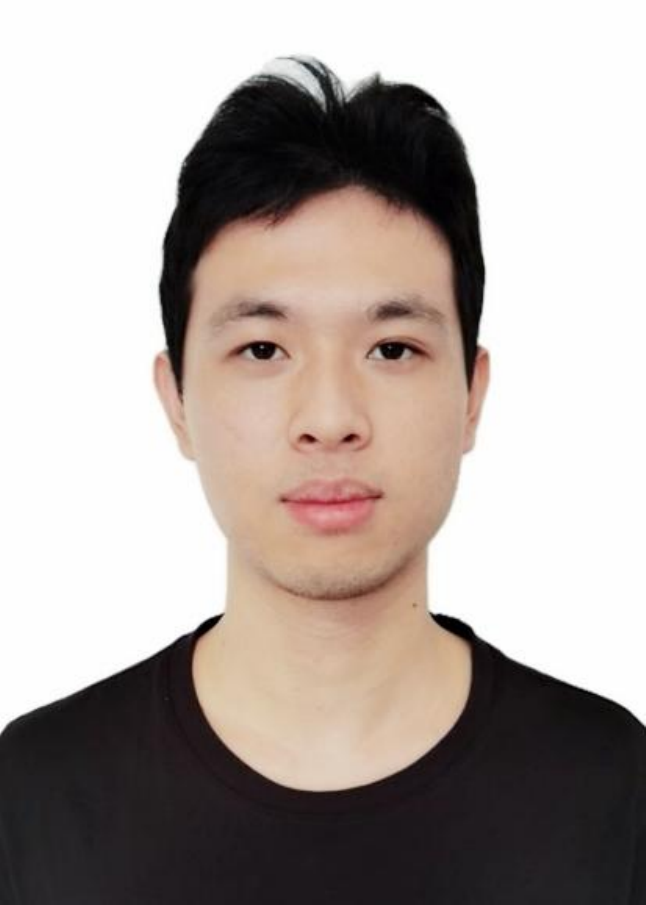}}]{Shixuan Liu}
received his B.S. and Ph.D. degrees from the National University of Defense Technology, Changsha, China, in 2019 and 2024, respectively. He is also a visiting scholar in the Department of Computer Science and Technology at Tsinghua University, where he has spent two years. He has published over 10 papers in prestigious journals and conferences, including T-PAMI, T-KDE, T-CYB, and ICDM, focusing on knowledge reasoning and data mining.
\end{IEEEbiography}

\vspace{-12mm}
\begin{IEEEbiography}[{\includegraphics[width=1in,height=1.25in,clip,keepaspectratio]{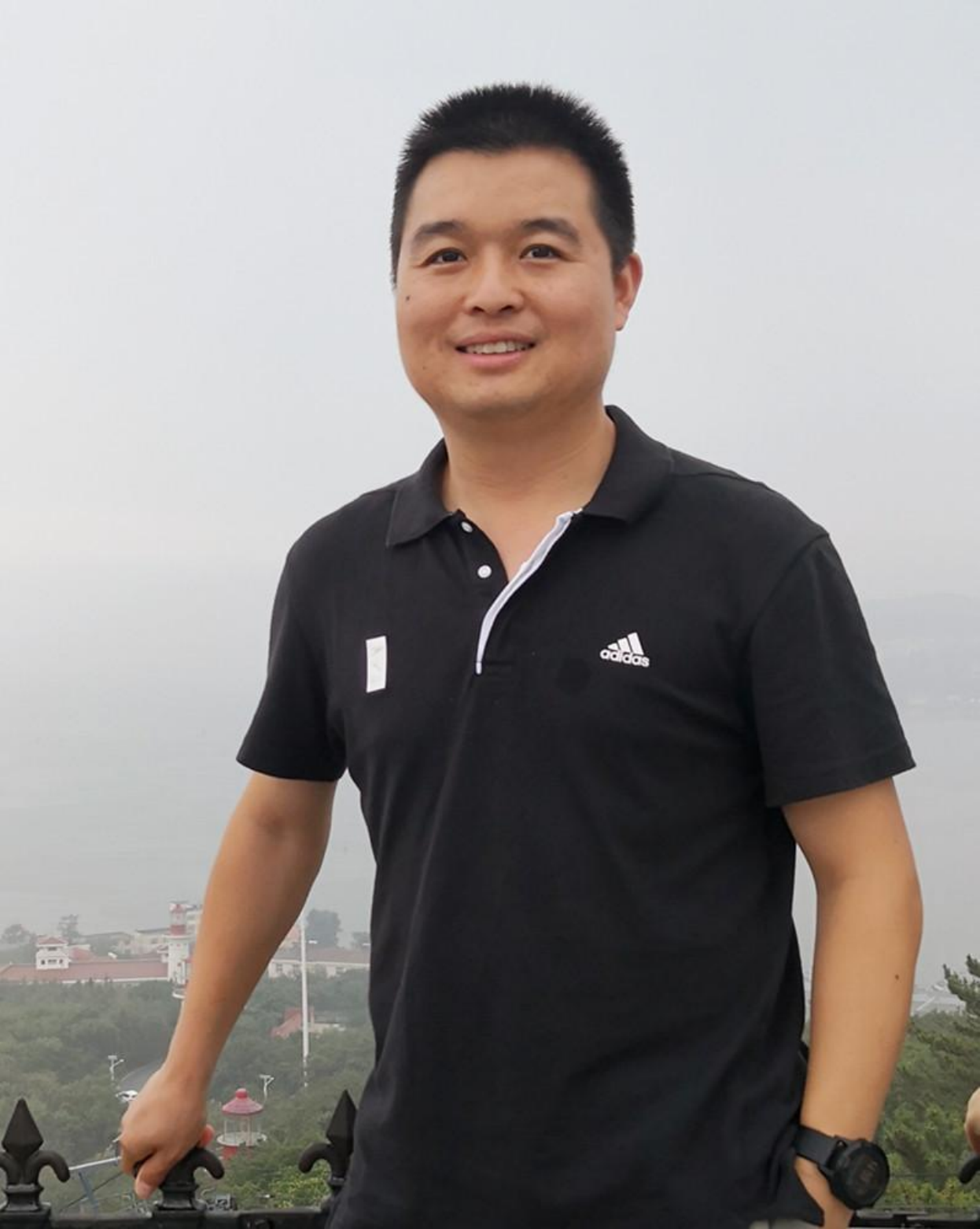}}]{Kuihua Huang}
received the Ph.D. degree from National University of Defense Technology, China, in 2015. He is currently an associate professor at National University of Defense Technology. His research interests include artificial intelligence, parallel simulation and intelligent optimization.
\end{IEEEbiography}

\vspace{-12mm}
\begin{IEEEbiography}[{\includegraphics[width=1in,height=1.25in,clip,keepaspectratio]{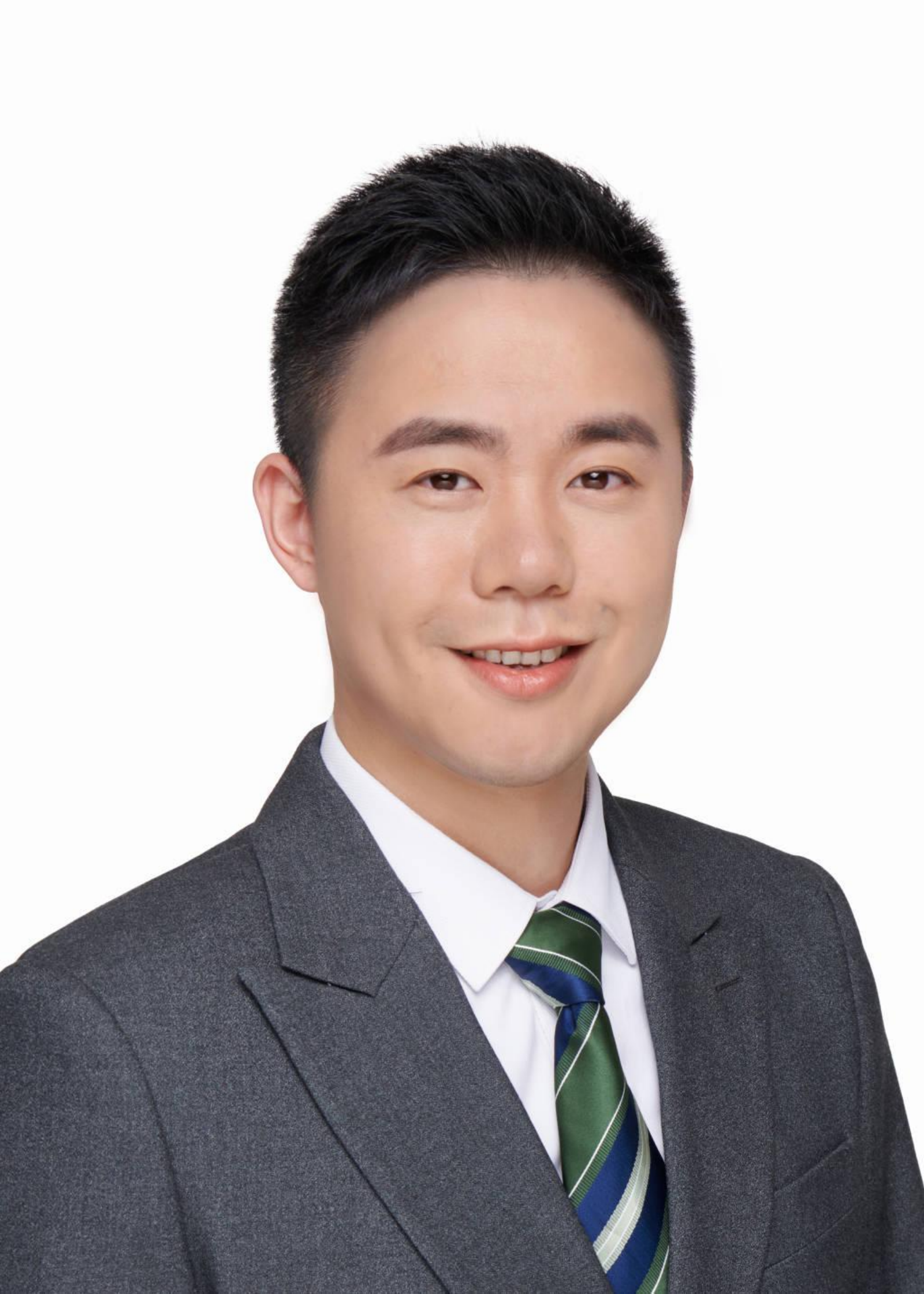}}]{Xiang Zhao}
received the PhD degree from The University of New South Wales, Australia, in 2014. He is currently a professor at National University of Defense Technology, China. His research interests include graph data management and mining, with a special focus on knowledge graphs.
\end{IEEEbiography}

\vspace{-15mm}
\begin{IEEEbiography}[{\includegraphics[width=1in,height=1.25in,clip,keepaspectratio]{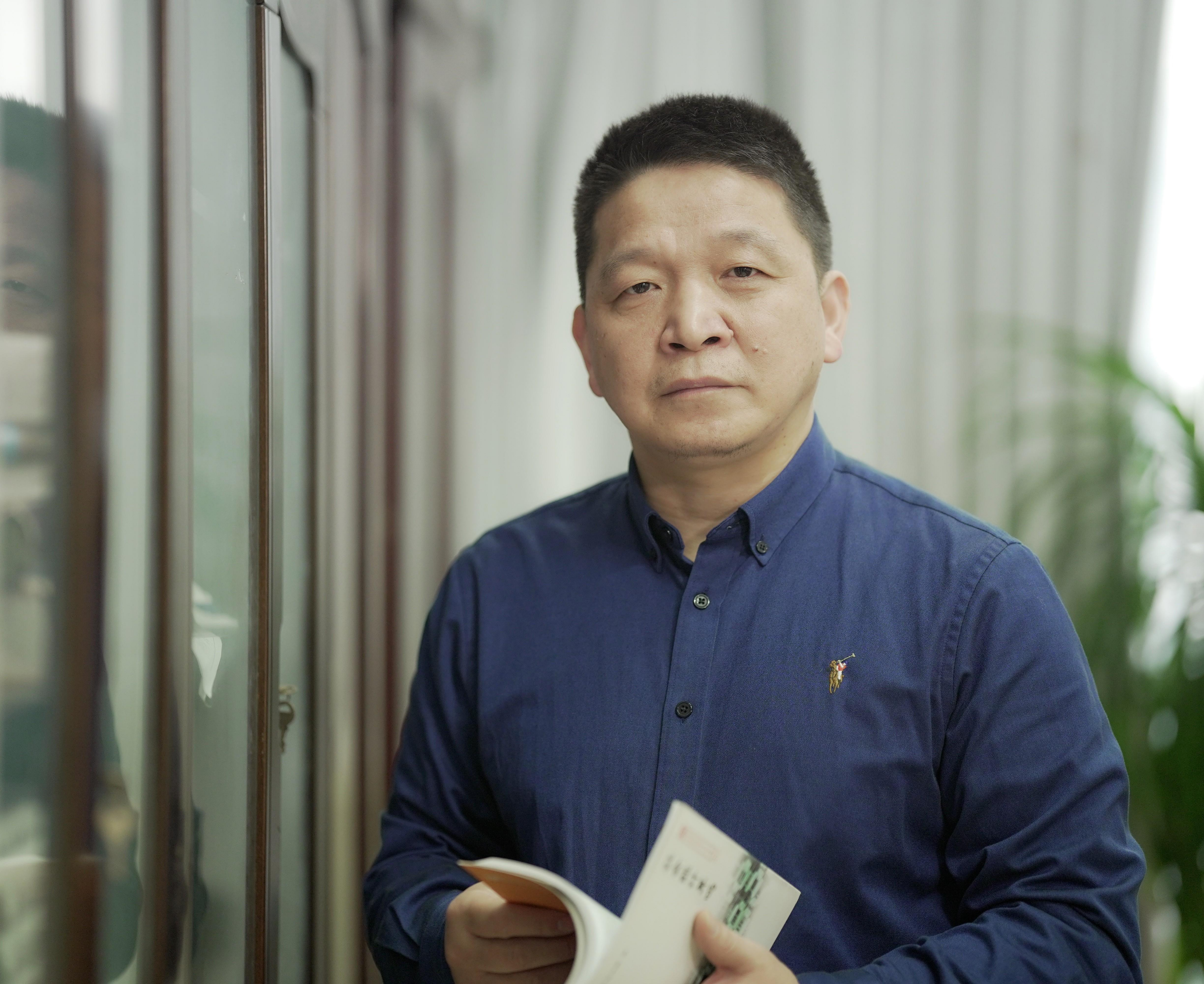}}]{Jincai Huang}
is currently a Professor of the National University of Defense Technology, Changsha, Hunan, China; and a Researcher of the Science and Technology on Information Systems Engineering Laboratory. His main research interests include artificial general intelligence, deep reinforcement learning, and multi-agent systems.
\end{IEEEbiography}

\vspace{-15mm}
\begin{IEEEbiography}[{\includegraphics[width=1in,height=1.25in,clip,keepaspectratio]{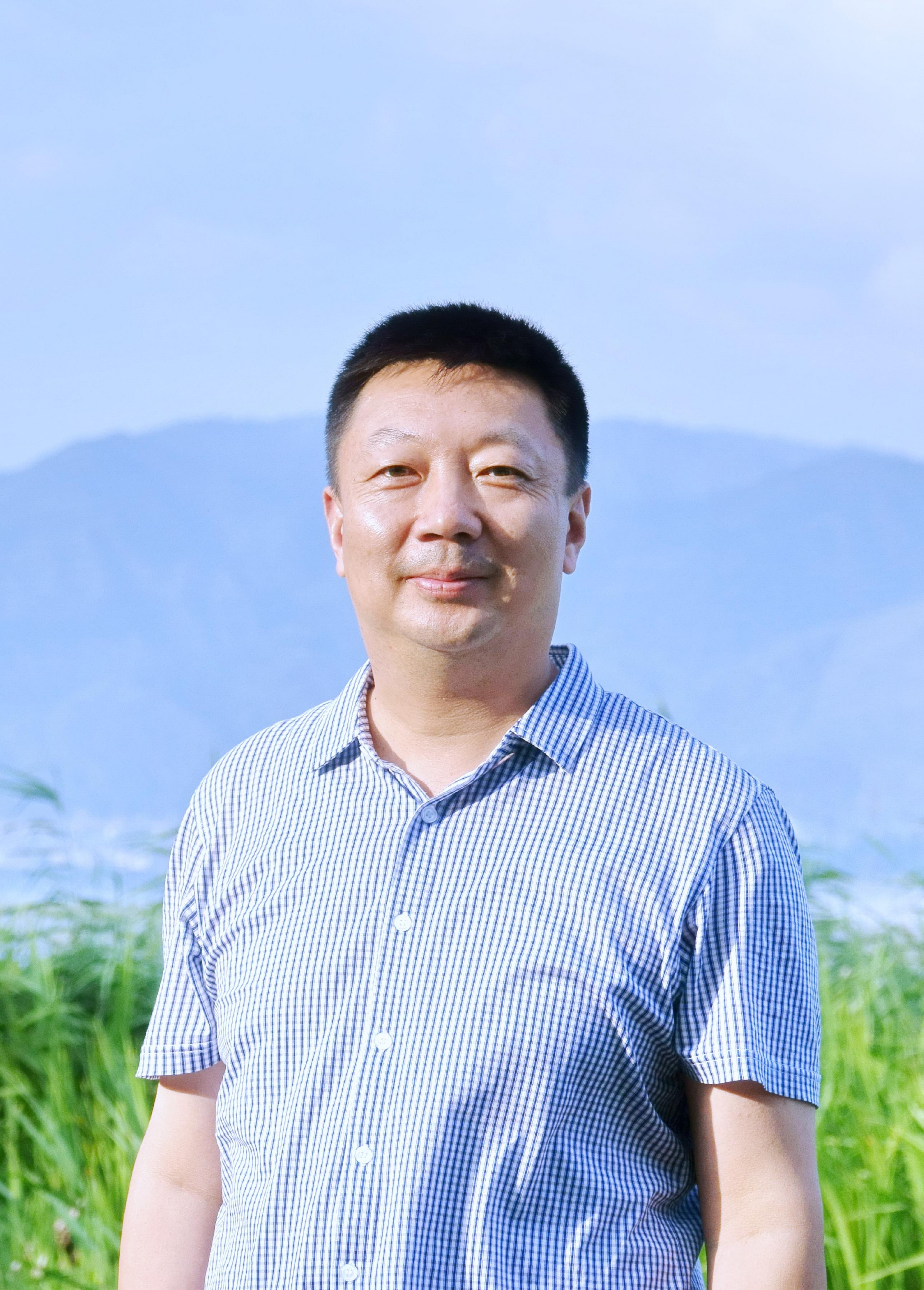}}]{Zhong Liu}
received the B.S. degree in Physics from Central China Normal University, Wuhan, Hubei, China, in 1990, the M.S. degree in computer software and the Ph.D. degree in management science and engineering both from National University of Defense Technology, Changsha, China, in 1997 and 2000. He is a professor in the College of Systems Engineering, National University of Defense Technology, Changsha, China. His research interests include intelligent information systems, and intelligent decision making.
\end{IEEEbiography}

\end{document}